%% file: example_paper.tex
\documentclass{article}

\usepackage{microtype}
\usepackage{graphicx}
\usepackage{subcaption}
\usepackage{booktabs} 

\usepackage{hyperref}
\input{math_commands.tex}

\input{init.tex}

\usepackage[accepted]{icml2026}

\usepackage{amsmath}
\usepackage{amssymb}
\usepackage{mathtools}
\usepackage{amsthm}

\usepackage[capitalize,noabbrev]{cleveref}

\theoremstyle{plain}

\theoremstyle{definition}

\theoremstyle{remark}

\usepackage[textsize=tiny]{todonotes}

\icmltitlerunning{\archcolor{}: Layerwise Timestep-Expert Flow-based Transformer}

\begin{document}

\twocolumn[
  \icmltitle{\TitleIcon\; \archcolor{}: Layerwise Timestep-Expert Flow-based Transformer}



  \icmlsetsymbol{equal}{*}
  \icmlsetsymbol{sup}{\dagger}

  \begin{icmlauthorlist}
    \icmlauthor{Ying Shen}{equal,uiuc}
    \icmlauthor{Zhiyang Xu}{equal,vt}
    \icmlauthor{Jiuhai Chen}{maryland}
    \icmlauthor{Shizhe Diao}{nvidia}
    \icmlauthor{Jiaxin Zhang}{saleforce}
    \icmlauthor{Yuguang Yao}{intuit}
    \icmlauthor{Joy Rimchala}{intuit}
    \icmlauthor{Ismini Lourentzou}{uiuc,sup}
    \icmlauthor{Lifu Huang}{davis,sup}
  \end{icmlauthorlist}

  \icmlaffiliation{uiuc}{University of Illinois Urbana-Champaign}
  \icmlaffiliation{vt}{Virginia Tech}
  \icmlaffiliation{maryland}{University of Maryland}
  \icmlaffiliation{davis}{UC Davis}
  \icmlaffiliation{nvidia}{Nvidia}
  \icmlaffiliation{saleforce}{Salesforce AI Research}
  \icmlaffiliation{intuit}{Intuit AI Research}

  \icmlcorrespondingauthor{Ying Shen}{ying22@illinois.edu}
  \icmlcorrespondingauthor{Zhiyang Xu}{zhiyangx@vt.edu}

  \icmlkeywords{Machine Learning, ICML}

  \vskip 0.3in
]


\printAffiliationsAndNotice{\icmlEqualContribution  \textsuperscript{\dagger}Equal supervision.}

\input{sections/0_abstract}
\input{sections/1_introduction}
\input{sections/2_related_work}

\input{sections/3_preliminary}

\input{sections/3_method}

 \input{sections/4_experiment_setup}

\input{sections/5_results_discussion}
\input{sections/6_conclusion}

\section*{Impact Statement}

This work advances the field of efficient unified multimodal modeling by introducing \arch{}, a novel architecture that improves the efficiency of diffusion/flow-based transformer within the unified model setting. By leveraging pretrained vision-language models and introducing novel architectural mechanisms, Layerwise Timestep Experts and Timestep-Conditioned Residual Attention, \arch{} achieves strong performance with significantly improved inference speed.

The proposed model has a potential impact in both academic and practical settings, as a scalable solution for building efficient, unified multimodal foundation models. 
This work has the potential to enable more efficient deployment of multimodal systems in resource-constrained environments, such as mobile devices or real-time applications, while maintaining high performance. 

While \arch{} improves performance and efficiency, it inherits the biases of its pretrained vision-language foundation and may generate misleading or inappropriate outputs if not properly constrained. Careful evaluation and mitigation of such risks are important for downstream deployment.

\section*{Acknowledgments}
 This research was partially supported by the National Science Foundation under CAREER Award (\#2604309, \#2542328), the Intuit Faculty Research Award, the Defense Advanced Research Projects Agency (DARPA) under award HR001125C0303, and the U.S. Army under contract W5170125CA160. Any opinions, findings, conclusions, or recommendations expressed in this material are those of the authors and do not necessarily reflect the views of NSF, DARPA, the U.S. Army, or the U.S. Government. The U.S. Government is authorized to reproduce and distribute reprints for governmental purposes notwithstanding any copyright annotation therein.
 
\bibliography{example_paper}
\bibliographystyle{icml2026}

\newpage
\appendix
\onecolumn

\input{sections/X_sup}

\end{document}

%% file: math_commands.tex
\usepackage{amsmath,amsfonts,bm}

\usepackage[normalem]{ulem}

\def\eqref#1{equation~\ref{#1}}

\def\1{\bm{1}}

\DeclareMathAlphabet{\mathsfit}{\encodingdefault}{\sfdefault}{m}{sl}
\SetMathAlphabet{\mathsfit}{bold}{\encodingdefault}{\sfdefault}{bx}{n}



%% file: init.tex
\usepackage{graphicx}               
\usepackage{tabularx}               
\usepackage{booktabs}
\usepackage{amsmath}
\usepackage{stmaryrd}

\usepackage{soul}

\newcolumntype{C}{>{\centering\arraybackslash}X}
\usepackage{multirow}               
\usepackage{diagbox}                
\usepackage{hhline}                 
\usepackage{color,colortbl}         
\usepackage{amsmath}                
\usepackage{amssymb}                
\usepackage{mathtools}              
\usepackage{enumitem}               

\usepackage{subcaption}
\usepackage{caption}
\usepackage{booktabs}
\usepackage{extarrows}
\usepackage{makecell}
\usepackage{hyperref}
\usepackage{fdsymbol}
\usepackage{graphicx}
\usepackage{tcolorbox}
\usepackage{relsize}
\usepackage{wrapfig}

\usepackage{amssymb}
\usepackage{pifont}

\definecolor{RoseQuartzBg}{HTML}{F7CAC9}
\definecolor{ForestGreen}{HTML}{228b22}
\definecolor{RoseQuartz}{HTML}{F5A798}
\definecolor{Serenity}{HTML}{92A8D1}
\definecolor{OrangeRed}{rgb}{1.0, 0.27, 0.0}
\definecolor{Red}{rgb}{1.0, 0.0, 0.0}
\definecolor{forestgreen}{rgb}{0.13, 0.55, 0.13}
\definecolor{Turquoise}{HTML}{0F4C81}
\definecolor{columbiablue}{rgb}{0.61, 0.87, 1.0}
\definecolor{Gray}{gray}{0.9}
\definecolor{lavenderPurple}{HTML}{7030A0}
\definecolor{methodblue}{HTML}{0070C0}

\usepackage{xparse}
\usepackage{footmisc}
\usepackage{fvextra}

\NewDocumentCommand{\ying}{ mO{} }{\textcolor{teal}{\textsuperscript{\textit{Ying}}\textsf{\textbf{\small[#1]}}}}

\NewDocumentCommand{\lifu}{ mO{} }{\textcolor{blue}{\textsuperscript{\textit{Lifu}}\textsf{\textbf{\small[#1]}}}}

\NewDocumentCommand{\zhiyang}{ mO{} }{\textcolor{Turquoise}{\textsuperscript{\textit{zhiyang}}\textsf{\textbf{\small[#1]}}}}

\definecolor{lightblue}{rgb}{0.0, 0.5, 1.0}

\definecolor{DarkCoffee}{HTML}{4B2E2B}
\definecolor{MediumCoffee}{HTML}{6F4E37}
\definecolor{LightCoffee}{HTML}{A67B5B}
\definecolor{Golden}{HTML}{D4A017}
\newcommand{\archcolor}[1]{{\textcolor{DarkCoffee}{L}\textcolor{DarkCoffee}{a}\textcolor{MediumCoffee}{T}\textcolor{MediumCoffee!80}{t}\textcolor{MediumCoffee}{E}-\textcolor{Golden}{F}\textcolor{Golden!90}{l}\textcolor{Golden!80}{o}\textcolor{Golden!70}{w}}}

\newcommand{\archlong}[1]{{Layerwise Timestep-Expert Flow-based Transformer}} %

\newcommand{\arch}[1]{{LaTtE-Flow}} 
\newcommand{\archlongcolor}[1]{{\textcolor{DarkCoffee}{La}yerwise \textcolor{MediumCoffee}{T}imes\textcolor{MediumCoffee}{t}ep-\textcolor{MediumCoffee}{E}xpert \textcolor{Golden}{F}\textcolor{Golden!90}{l}\textcolor{Golden!80}{o}\textcolor{Golden!70}{w}-based 
Transformer}} 

\newcommand{\blend}[1]{{Blend}} 
\newcommand{\couple}[1]{{Couple}} 

\newcommand{\baselineblend}[1]{{Vanilla Blend}} 
\newcommand{\baselinecouple}[1]{{Vanilla}} 

\newcommand\TitleIcon{%
  {\raisebox{-0.26\baselineskip}{
    \includegraphics[height=1.2em]{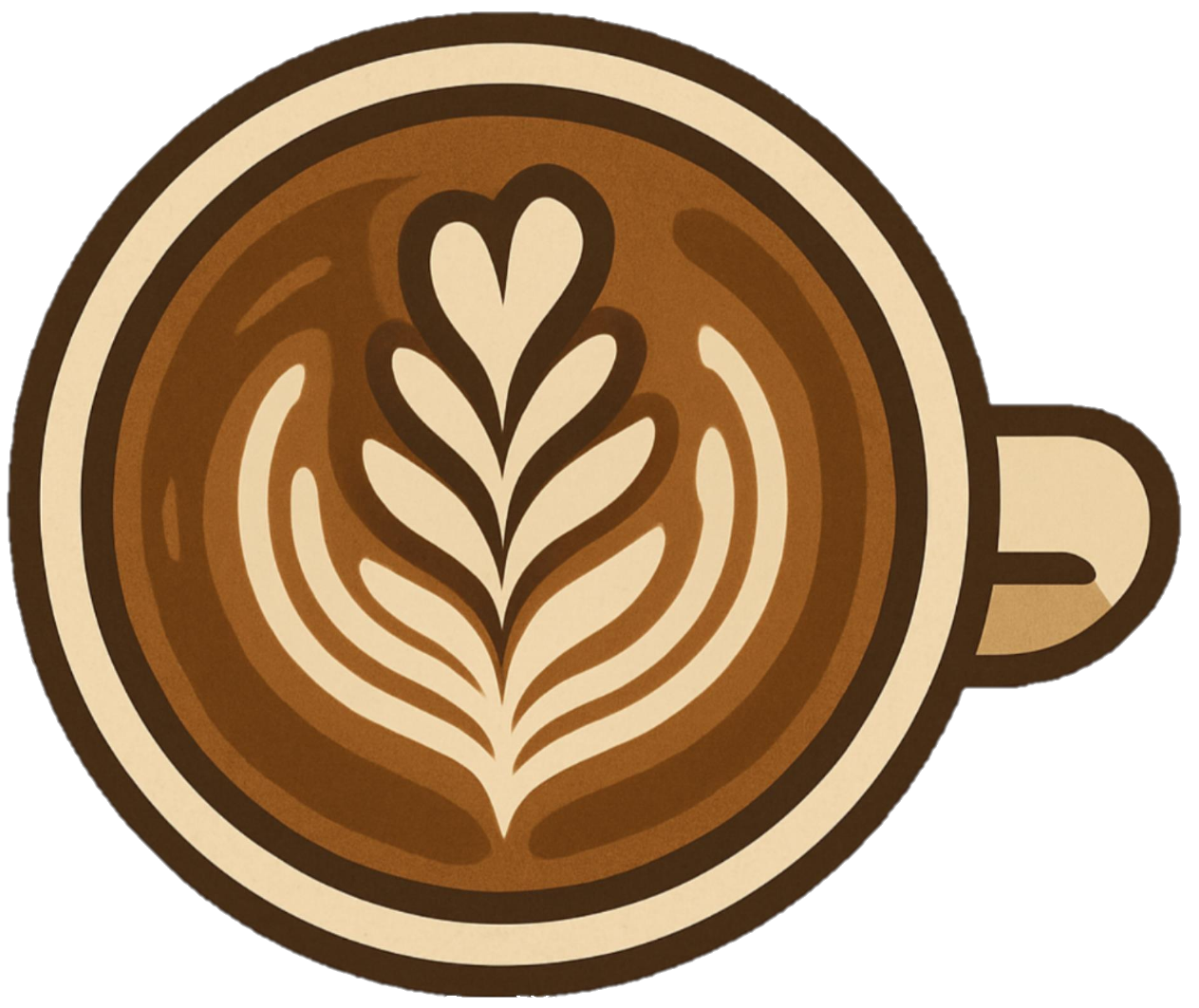}
  }
  }%
}

\definecolor{Espresso}{HTML}{4B2E2B}
\definecolor{Mocha}{HTML}{6F4E37}
\definecolor{Latte}{HTML}{A67B5B}
\definecolor{Cappuccino}{HTML}{D2B48C}
\definecolor{Caramel}{HTML}{C08051}
\definecolor{Cream}{HTML}{F5F5DC}

%% file: sections/0_abstract.tex
\begin{abstract}
Recent advances in multimodal foundation models unifying image understanding and generation have opened exciting avenues for tackling a wide range of vision-language tasks within a single framework. 
Despite progress, existing unified models often rely on extensive pretraining and suffer from slow generation speeds, limiting their practical deployment in real-time and resource-constrained settings.
In this work, we introduce \textbf{\archlongcolor{}} (\textbf{\archcolor{}}), a novel architecture that improves the efficiency of diffusion/flow-based Transformers within the unified model setting.
\arch{} builds upon powerful pretrained Vision-Language Models (VLMs) to inherit strong multimodal understanding capabilities, and extends them with a novel Layerwise Timestep Experts flow-based architecture for efficient image generation.
\arch{} distributes the flow-matching process across specialized groups of Transformer layers, each responsible for a distinct subset of timesteps.
This design significantly improves sampling efficiency by activating only a small subset of layers at each sampling timestep.
To further enhance performance, we propose a Timestep-Conditioned Residual Attention mechanism for efficient information reuse across layers.
Experiments demonstrate that \arch{} achieves strong performance on multimodal understanding tasks, while achieving competitive image generation quality with around \textbf{6}$\times$ faster inference speed compared to recent unified multimodal models.

\end{abstract}

%% file: sections/1_introduction.tex
\section{Introduction}

Recent advances in multimodal foundation models capable of 
both image understanding and generation have opened promising avenues for building unified architectures that support a wide range of vision-language tasks~\citep{shi2024llamafusion,wang2024emu3,xie2024show,zhou2024transfusion,chen2025janus,ma2024janusflow,tong2024metamorph}.
Such unified multimodal models hold great potential for building general-purpose agents that can interpret, reason about, and generate multimodal content in response to user instructions.
Current approaches to unified multimodal modeling generally fall into two broad categories.
The first category leverages vector-quantized autoencoders~\citep{van2017neural,esser2021taming,yu2021vector} to discretize images into token sequences, which are then incorporated into the vocabulary of Large Language Models (LLMs)~\citep{sun2024generative,wang2024emu3,xie2024show,wu2024janus,chen2025janus,wu2024vila}. 
These models are subsequently trained to autoregressively generate the next token, either textual or visual, thus integrating vision and language generation within a single framework. 
The second category leverages diffusion-based methods, either by coupling LLMs with external diffusion modules or by training LLMs to directly perform denoising steps~\citep{zhou2024transfusion,shi2024llamafusion,ma2024janusflow,tong2024metamorph,ge2024seed}.

Despite significant progress, existing unified multimodal models still require extensive pretraining and struggle to support both multimodal understanding and image generation in an effective and efficient manner within a single architecture~\citep{shen2025efficient,xiong2024autoregressive}.
For example, unified models that leverage diffusion or flow-matching processes require dozens of forward passes through the full backbone during inference, resulting in slow and resource-intensive generation~\citep{shen2025efficient}. 
Similarly, autoregressive approaches suffer from long decoding times, especially for high-resolution images that require generating large numbers of tokens sequentially~\citep{xiong2024autoregressive}.

\input{figures/diffusion_process}

To address these challenges, we propose \textbf{\archlongcolor{} (\archcolor{})}, 
a novel architecture that improves the efficiency of diffusion/flow-based transformer within the unified model setting.
In particular, \arch{} builds upon existing pre-trained VLMs that already possess strong multimodal understanding capabilities, and further introduces two key architectural innovations designed to enable efficient and high-quality image generation.
First, we propose a novel \textbf{\textcolor{DarkCoffee}{Layerwise Timestep Expert architecture}}, which reduces the sampling time complexity by distributing the flow-matching process across groups of transformer layers. 
Instead of invoking the entire model across all time steps, \arch{} partitions transformer layers into disjoint groups, each assigned to a specific range of timesteps in the flow-matching process, as shown in Figure \ref{fig:diffusion_process}.
During inference, only the relevant expert group is activated at each timestep, which drastically reduces computation while preserving generation quality. 
Second, we introduce \textbf{\textcolor{MediumCoffee}{Timestep-Conditioned Residual Attention}}, a lightweight mechanism that enables later layers to reuse self-attention maps computed at earlier layers, modulated by the current timestep. This design encourages the model to gradually refine features across layers, resulting in faster convergence during training. 
Experiments demonstrate that these two innovations enable \arch{} to achieve efficient and high-quality image generation. For example, \arch{} attains competitive generation quality with around 6$\times$ faster inference compared to recent unified models on ImageNet~\cite{imagenet}, while maintaining strong multimodal understanding performance across several benchmark datasets. Extensive ablation studies highlight that \arch{} accelerates convergence and inference while preserving strong generation quality.

In summary, our contributions are:
\textbf{(1)} We propose \arch{}, an efficient unified multimodal architecture that integrates flow-matching-based image generation with pre-trained vision-language models.
\textbf{(2)} We introduce the Layerwise Timestep Expert architecture, a novel design that significantly reduces inference complexity by distributing transformer layers into timestep-specific experts.
\textbf{(3)} We design a Timestep-Conditioned Residual Attention module, which enables effective reuse of attention information across layers, boosting training efficiency and performance. 
\textbf{(4)} Extensive experiments demonstrate that \arch{} achieves competitive performance on both generation and understanding tasks, while offering 6$\times$ faster inference compared to recent unified models.

%% file: figures/diffusion_process.tex
\begin{figure*}[t!]
	\centering
	\includegraphics[width=0.99\linewidth, trim={0 0 0 0},clip ]{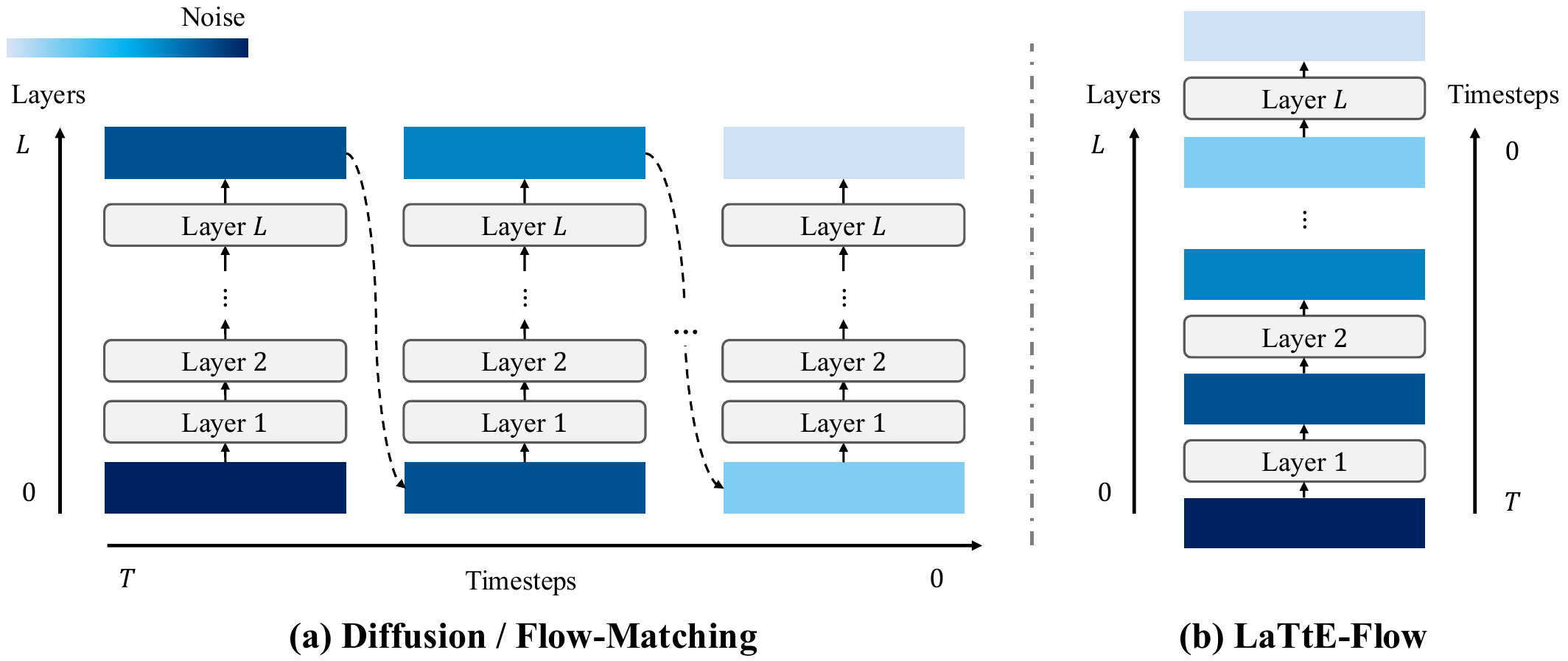}
    \vspace{-0.2cm}
	\caption{\textbf{Comparison between standard diffusion / flow-matching vs. our proposed \arch{}.} Unlike diffusion / flow-matching based models, which invoke the entire model at each sampling timestep, \arch{} activates only a subset of layers at each step, improving efficiency. }
\label{fig:diffusion_process}
\vspace{-0.2cm}
\end{figure*}

%% file: sections/2_related_work.tex
\section{Related Work}

\paragraph{Unified Models.}

Unified multimodal architectures integrate multimodal understanding and generation within a single model, enabling 
general-purpose agents that can interpret and generate multimodal content in response to user instructions~\citep{shi2024llamafusion,wang2024emu3,xie2024show,zhou2024transfusion,chen2025janus,ma2024janusflow,tong2024metamorph}.
Existing approaches to unified modeling primarily fall into two categories:
The first class of models relies on vector-quantized autoencoders~\cite{van2017neural,esser2021taming,yu2021vector} to convert images into discrete token sequences that can be processed similarly to text. These visual tokens are added to the LLM vocabulary to enable unified autoregressive training over both language and vision~\citep{sun2024generative,wang2024emu3,xie2024show,wu2024janus,chen2025janus,wu2024vila}. 
The second class incorporates continuous generative processes, most notably diffusion models~\citep{ho2020denoising} or flow-matching models~\citep{lipmanflow}. Some approaches connect LLMs with external diffusion modules, using the language model to guide image generation~\citep{tong2024metamorph,ge2024seed,metaqueries,chen2025blip3ofamilyfullyopen,mossXu2025}, while others directly train LLMs to jointly perform denoising or flow-matching steps~\citep{zhou2024transfusion,shi2024llamafusion,ma2024janusflow}. 
Despite progress in both categories, many of these models suffer from slow image generation speeds, limiting their practical deployment in resource-constrained settings.

\paragraph{Multiple Experts in Diffusion Models.}

Recent advancements in diffusion models have increasingly adopted modular or expert-based architectures for better image generation~\cite{sun2024ec,shi2025diffmoe}. 
Building on this direction, several recent approaches have explored the use of expert models tailored to different diffusion timesteps~\citep{lee2024multi,fang2024remix,zhuang2025timestep}.
By allocating distinct experts to specific temporal intervals, these models aim to better capture the evolving nature of the denoising process. This design is partly motivated by findings from prior work~\cite{HangGLB00GG23,eDiff}, which show that optimization gradients from different timesteps often conflict, leading to slower convergence and degraded model performance. 
However, these models typically maintain a near full-parameter expert network for different timestep intervals, which leads to little or no improvement in inference efficiency under a fixed number of sampling steps. 
In contrast, we introduce a layerwise timestep expert architecture, which partitions the transformer layers into different groups of layers, each responsible for a specific range of timesteps. At inference time, only the corresponding group is activated, significantly reducing the number of parameters involved at each step. 
Moreover, our design allows all expert groups to be trained jointly, and we further integrate it within a unified model architecture, enhancing both efficiency and performance.

%% file: sections/3_preliminary.tex
\section{Preliminaries}

\paragraph{Flow-Matching.}
Flow-based generative models~\citep{lipmanflow,liuflow,albergobuilding} aim to learn a time-dependent velocity field $\bm{v}_t$ that transports samples from a simple source distribution $p_0(\bm{x})$ (e.g., standard Gaussian) to a complex target distribution $p_1(\bm{x})$ via an ordinary differential equation (ODE):
\begin{align}
    \frac{d \bm{x}_t}{d t} = \bm{v}_t (\bm{x}_t), \quad  \bm{x}_0 \sim p_0(\bm{x}).
    \label{eq:ode}
\end{align}
Recently, \citet{lipmanflow} propose a simple simulation-free Conditional Flow Matching (CFM) objective by defining a conditional probability path $p_t(\bm{x}_t \mid \bm{x}_1)$ and the corresponding conditional vector field $\bm{u}_t(\bm{x}_t\mid\bm{x}_1)$ per sample $\bm{x}_1$.
The model directly regresses the velocity $\bm{v}_t$ on a conditional vector field $\bm{u}_t(\cdot \mid \bm{x}_1)$:
\begin{align}
    \mathbb{E}_{t, p_1(\bm{x}_1), p_t(\bm{x}_t \mid \bm{x}_1)} \| \bm{v}_t(\bm{x}_t, t) - \bm{u}_t(\bm{x}_t \mid \bm{x}_1) \|^2,
\end{align}
where $\bm{u}_t(\cdot \mid \bm{x}_1)$ uniquely determines a conditional probability path $p_t(\cdot \mid \bm{x}_1)$ towards target data sample $\bm{x}_1$. 
A widely adopted choice for the conditional probability path is linear interpolation between the source and target data~\citep{liuflow}: $\bm{x}_t\!=\!t \bm{x}_1 + (1-t) \bm{x}_0$. 
Assuming the source distribution $p_0$ is a standard Gaussian, this yields $\bm{x}_t\!\sim\!\mathcal{N}(t \bm{x}_1, (1-t)^2 \bm{I})$. 
Sampling from the learned model is obtained by sampling $\bm{x}_0\!\sim\!\mathcal{N}(\bm{x} \mid 0, 1)$ and then numerically solving the ODE in Eq. (\ref{eq:ode}).

%% file: sections/3_method.tex
\section{\textbf{\archcolor{}}}
We present \arch{} (\archlong{}), a novel architecture designed for efficient and high-quality image generation and multimodal understanding, unified within a single model. 
Built on top of pretrained Vision-Language Models (VLMs), \arch{} leverages their powerful understanding capabilities while introducing additional flow-matching based generation components to enable scalable and effective image synthesis.
As illustrated in Figure \ref{fig:diff_vlm_arch}, \arch{} is implemented as a mixture-of-transformer architecture, allowing for effective interaction between image latents and multimodal context.
We also explore alternative architecture variants using a single transformer as the backbone in Appendix~\ref{app:blend}, to highlight that our proposed method is not restricted to a single form.

Furthermore, we introduce two core architectural innovations applicable to both variants to enhance image generation efficiency and quality:
\textbf{(1) Layerwise Timestep Experts} (Section \ref{sec:layerwise}), which partition the model into timestep-specialized modules to reduce sampling complexity, and 
\textbf{(2) Timestep-Conditioned Residual Attention} (Section \ref{sec:attn}), which injects timestep-aware residual attention into each attention layer through gating mechanisms modulated by a learned timestep embedding, improving training efficiency through effective information reuse across layers.

\subsection{\arch{} Layer Design }
\label{sec:design}

\arch{} preserves the pretrained VLM \input{figures/overview} entirely, keeping its parameters frozen (shown in \textbf{\textcolor{lavenderPurple}{purple}} in Figure~\ref{fig:diff_vlm_arch}) to retain strong multimodal understanding without finetuning.
To enable image generation, it introduces a trainable generative pathway alongside the frozen backbone. Specifically, each Transformer layer is augmented with a trainable replica of the original VLM layer, along with additional components for flow-matching-based generation (shown in \textbf{\textcolor{methodblue}{blue}} in Figure~\ref{fig:diff_vlm_arch}). 
\arch{} thus allows the model to perform image synthesis while leveraging the robust understanding capabilities of the pretrained VLM.

As illustrated in Figure~\ref{fig:diff_vlm_arch}, we introduce a \arch{} Attention module to enable effective interaction between generative image latents and multimodal context.

Specifically, the noisy image latents, used during the flow-based generation process, attend to the text and visual context tokens, as detailed in Appendix~\ref{app:latte_attn}.
This attention module employs a hybrid positional encoding scheme, combining the original 3D Rotary Positional Embeddings (RoPE)~\citep{su2024roformer}, inherited from the pretrained VLM, for encoding spatial and temporal structure in the multimodal context, with newly introduced 2D positional encodings applied to the generative image tokens.

\subsection{Layerwise Timestep Experts}
\label{sec:layerwise}

Typical sampling procedures in diffusion models~\citep{song2019generative,ho2020denoising} or flow-matching models~\citep{lipmanflow,liuflow,albergobuilding} require repeatedly invoking the full network across a large number of timesteps, leading to slow inference-time speed. 
For instance, consider a standard diffusion transformer (DiT) model~\citep{peebles2023scalable} with $L$ transformer layers. The effective computational cost for $T$ sampling steps is $\mathcal{O}(L \times T)$, as shown in Figure~\ref{fig:diffusion_process} (a).
To alleviate this inefficiency, we introduce a novel Layerwise Timestep Expert architecture, which reduces the effective sampling time complexity by distributing the flow-matching process across groups of transformer layers.

Specifically, instead of executing the entire model at every timestep, we partition the $L$ transformer layers into $K$ non-overlapping groups, where each group specializes in denoising samples within a specific timestep interval, as illustrated in Figure \ref{fig:diffusion_process} (b).
This design effectively enables efficient sampling, as only a subset of the network needs to be executed at each timestep.

Let each expert group be denoted as $\mathcal{G}^{l,l+M}_k\!=\!\{l, l\!+\!1, \dots, l\!+\!M\}$, consisting of  $M\!=\!{L}/{K}$ consecutive layers (from layer $l$ to layer $l+M$).
During training, each layer group learns to predict the velocity field over its assigned timestep interval $[t_k, t_{k+1}]$ using a layerwise flow-matching loss.
Specifically, each layer group $\mathcal{G}^{l,l+M}_k$ receives the noisy latent image $\bm{x}_t \in \mathbb{R}^{N_x \times d}$ along with the multimodal context $\bm{m}^l$, derived from the preceding layer $l\!-\!1$, and predicts the velocity field $\bm{s}_\theta(\bm{x}_t, \bm{m}^l, t)$. 
Formally, for timestep $t \in [t_k, t_{k+1}]$, the layerwise flow-matching loss is defined as:
\begin{equation}
\resizebox{0.95\linewidth}{!}{$
\displaystyle
\begin{aligned}
\mathcal{L}_t
= &\ \mathbb{E}_{t, p_1(\bm{x}_1), p_t(\bm{x}_t \mid \bm{x}_1)}
\left\|
\mathcal{G}^{l,l+M}_k(\bm{x}_t, \bm{m}^l, t)
-
\bm{u}_t(\bm{x}_t \mid \bm{x}_1)
\right\|^2  \\
&\quad \textrm{for } t \in [t_k, t_{k+1}] .
\end{aligned}
$}
\label{eq:layerwise}
\end{equation}
where $\mathcal{G}^{l,l+M}_k(\cdot)$ denotes the prediction produced by the expert group and $\bm{u}_t(\bm{x}_t \mid \bm{x}_1)$ is the ground-truth velocity at timestep $t$.
By training each group exclusively on its respective timestep interval, \arch{} encourages timestep specialization, allowing the model to learn timestep-specific representations across the flow-matching process.

\textbf{Inference.} 
Let $C_{\text{layer}}$ denote the average forward compute cost of one Transformer layer per step. At inference time with $T'$ sampling steps, for each timestep $t \in [t_k, t_{k+1}]$, \arch{} activates only the associated expert layer group $\mathcal{G}^{l,l+M}_k$ to perform a forward pass from layer $l$ to layer $l+M$.
This process is repeated across all $T'$ timesteps, with only $M\!=\!L/K$ layers evaluated per step. 
The multimodal hidden states, required for conditioning at each transformer layer, are computed once at the start of the inference and cached for reuse across all timesteps. Given one-time caching cost $C_{\text{cache}}$, the total inference cost for \arch{} is $C_{\text{cache}} + T' \times M \times C_{\text{layer}}$.
In contrast, conventional diffusion models or flow-matching models execute all $L$ layers at every step, with total inference cost $C_{\text{cache}} + T' \times L \times C_{\text{layer}}$. The resulting relative speedup $S$ is
\begin{align}
    S &= \frac{\mathcal{C}_{\textrm{baseline}}}{\mathcal{C}_{\textrm{\arch{}}}} = \frac{C_{\text{cache}}+T' \times L \times C_{\text{layer}}}{C_{\text{cache}} + T' \times (L / K) \times C_{\text{layer}}} \nonumber \\
    &= \frac{K+\theta}{1 + \theta}, \quad \text{where~} \theta\!=\!\frac{C_{\text{cache}}}{T' \times M \times C_{\text{layer}}}.
    \label{eq:pointless}
\end{align}
Since the one-time cache cost $C_{\text{cache}}$ is typically negligible compared to the cumulative compute across all sampling iterations $T'$.
As the number of sampling steps $T'$ grows, the one-time cache cost is amortized, i.e., $\theta\!\rightarrow\!0$ and hence $S\!\rightarrow\!K$. The resulting speedup shows that \arch{} guarantees a $K$-fold reduction in per-step compute cost, and a complexity reduction from $\mathcal{O}(L \times T')$ to $\mathcal{O}(M \times T')$.

\subsection{Timestep-Conditioned Residual Attention}
\label{sec:attn}

\input{figures/res_attn_fig}
To facilitate information reuse across transformer layers and improve both training efficiency and generative performance, we propose Timestep-Conditioned Residual Attention, a novel mechanism that introduces adaptive residual connections between successive image attention layers based on the current timestep.
Inspired by the success of residual connections in ResNet~\citep{he2016deep}, this design allows later layers to reuse and refine the attention patterns computed in earlier layers, while dynamically controlling the influence of past attention through the current flow-matching timestep.

Let $\bm{A}^l \in \mathbb{R}^{H \times N_x \times N_x}$ image self-attention matrix at layer $l$, where $N_x$ is the number of image tokens. In a standard self-attention layer, the attention matrix is computed as:
\begin{equation}
    \bm{A} = \mathrm{Softmax}\left(\frac{(\bm{h} \bm{W}^Q) (\bm{h} \bm{W}^K)^T}{\sqrt{d}}\right),
\end{equation}
where $\bm{h} \in \mathbb{R}^{N_x \times d}$ denotes the hidden states of the noisy image latents, and $\bm{W}^Q, \bm{W}^K \in \mathbb{R}^{d \times d}$ are learnable query and key projection matrices.
To incorporate residual attention from the previous layer, we define the augmented self-attention matrix at layer $l+1$ as:
 \begin{align}
     \tilde{\bm{A}}^{l+1} = \bm{A}^{l+1} + g(t) \odot \bm{A}^{l}, \quad  g(t) = \tanh(\bm{h}_t \bm{W}_t),
 \end{align}
 where $\bm{h}_t \in \mathbb{R}^{d}$ is the embedding of the current flow-matching timestep $t$ and $\bm{W}_t \in \mathbb{R}^{d \times H}$ is a trainable projection matrix, with $d$ denoting the hidden dimension and $H$ the number of attention heads. 
 The head-wise gating vector $g(t) \in (-1, 1)^H$, produced by a $\tanh(\cdot)$ activation, dynamically controls the extent to which each attention head incorporates residual attention information from the previous layer. The operator $\odot$ denotes element-wise multiplication, broadcast across all attention heads.
 
Notably, while the \arch{} Attention module jointly processes both noisy image states and multimodal hidden states, the residual attention mechanism is applied only to the self-attention map over the noisy image hidden states, as shown in Figure \ref{fig:res_attn}.
The timestep-conditioned residual attention mechanism enables the model to dynamically control how much residual attention from the previous layer is incorporated into the current layer, on a per-head basis and conditioned on the timestep.
Empirically, this design accelerates convergence during training and enhances the quality of generated images.

%% file: figures/overview.tex
\begin{figure}[!t]
	\centering
	\includegraphics[width=0.99\linewidth]{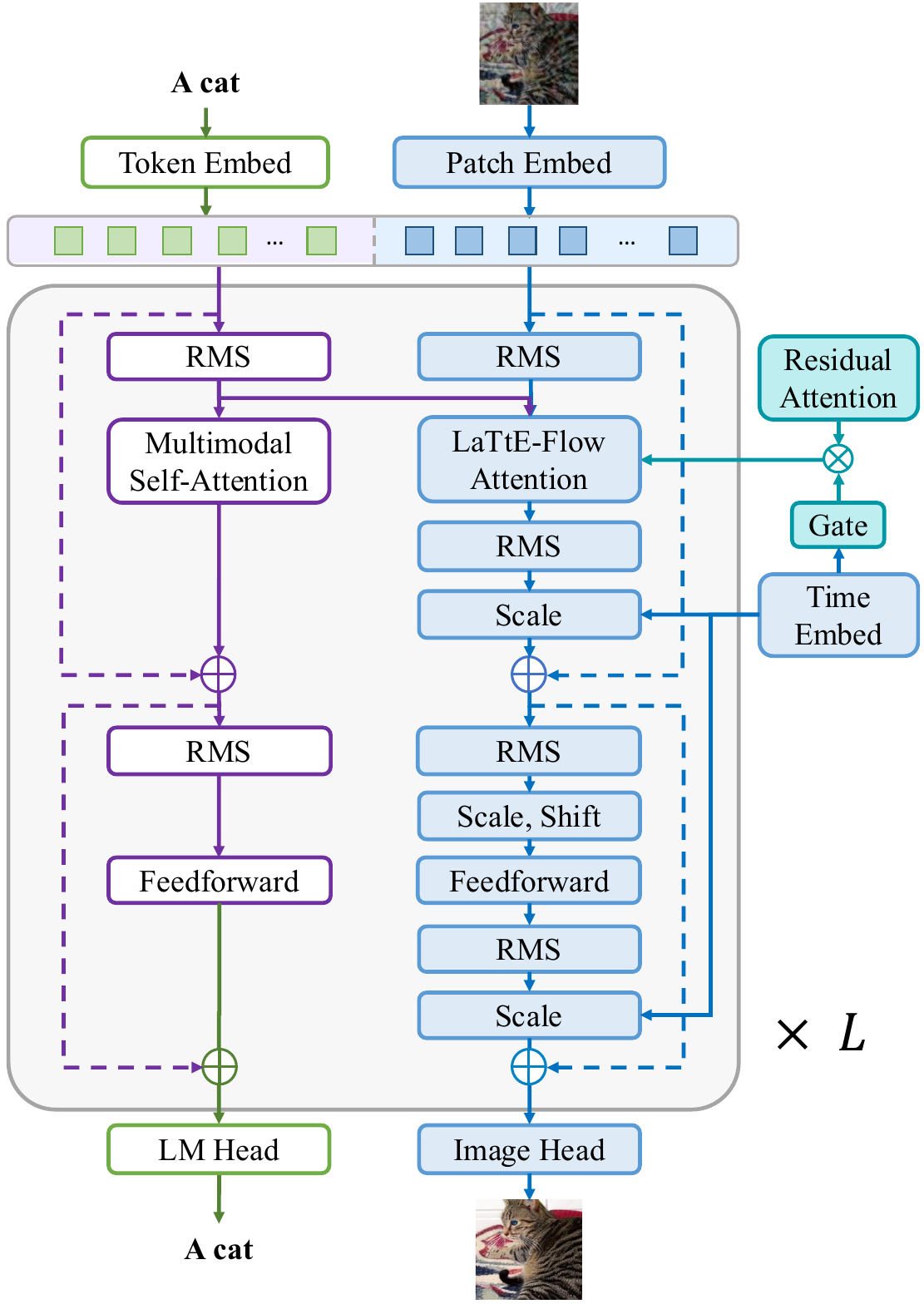}
	\caption{\textbf{\archcolor{} overall architecture}. 
    }
\label{fig:diff_vlm_arch}
\vspace{-0.5cm}
\end{figure}

%% file: figures/res_attn_fig.tex
\begin{figure}[!t] 
    \centering
\includegraphics[width=0.9\linewidth]{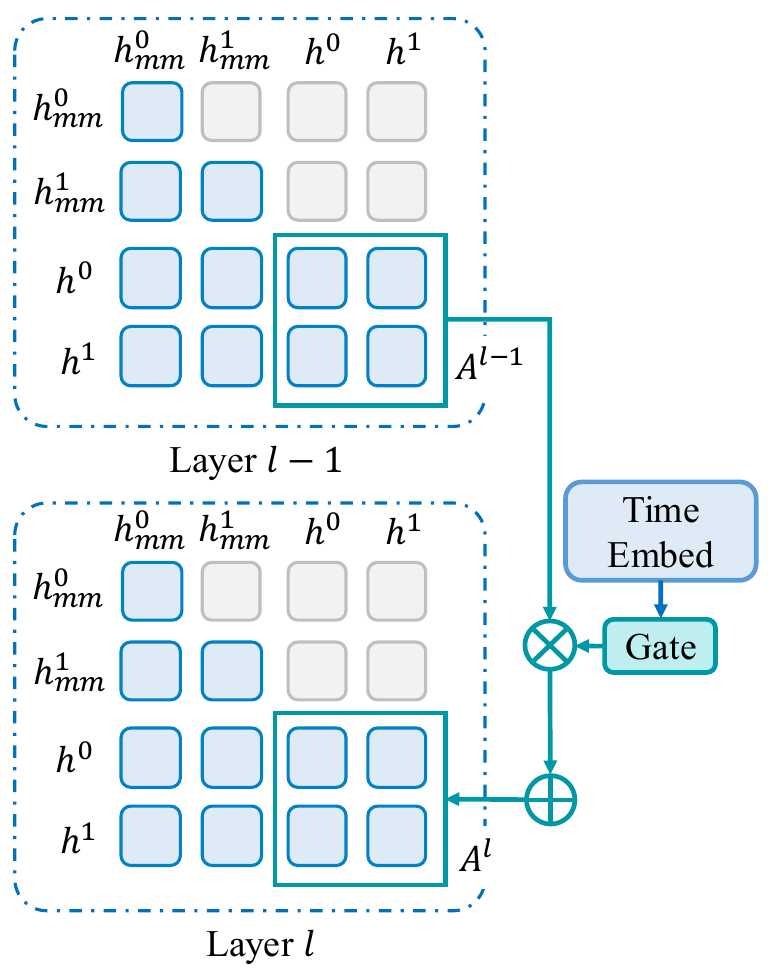}
    \vspace{-0.3cm}
    \caption{\textbf{Timestep-conditioned} \textbf{residual attention}}
    \label{fig:res_attn}
    \vspace{-0.4cm}
\end{figure}

%% file: sections/4_experiment_setup.tex
\section{Experiment Setup}
\label{sec:exp}
\textbf{Backbone Model and Image Encoder.} 
\arch{} is built upon Qwen2-VL-2B-Instruct~\citep{wang2024qwen2}, a pretrained VLM composed of $L\!=\!28$ transformer layers. 
 We create a trainable copy of each Transformer layer from the original Qwen2-VL-2B-Instruct and integrate it with additional components tailored for flow-matching-based image generation. These duplicated components are initialized with the corresponding pretrained weights from the original VLM.
For image encoding, we adopt the recently proposed Deep Compression Autoencoder (DC-AE)~\cite{dcae}, which compresses raw image pixels into a compact latent space using a $32\times$ down-sampling ratio. 

\input{tables/gen_results}

\textbf{Timestep Distribution.}
To enable Layerwise Timestep Experts, \arch{} partitions the model into $K\!=\!4$ non-overlapping layer groups, each containing $M\!=\!7$ consecutive layers for the final results. These groups are designed to operate over distinct intervals of the flow-matching timesteps.
During training, we use $T\!=\!1000$ flow-matching steps, which are initially divided uniformly into four intervals. To encourage robustness near interval boundaries and promote smooth transitions across groups, we introduce a 100-step overlap between adjacent timestep intervals during training. This overlap allows boundary timesteps to be seen by multiple layer groups, improving generalization.
At inference time, we disable the overlaps to maintain strict partitioning of timestep intervals.
Consequently, at each denoising step, only the corresponding expert layer group is activated, requiring just $M\!=\!7$ layers per inference step. This contrasts favorably with standard diffusion or flow-matching models that activate all $L\!=\!28$ layers at every step, significantly enhancing generation efficiency.
Further details are provided in Appendix \ref{app:timestep}.

\input{tables/mu_results}
\textbf{Baseline Architectures.}
We construct the baseline model  \baselinecouple{}, which matches the architecture of \arch{}, but excludes both the Layerwise Timestep Experts and Timestep-Conditioned Residual Attention mechanisms, allowing us to directly evaluate the effectiveness of these proposed mechanisms.
The \baselinecouple{} baseline retains a parallel generative path alongside the original VLM modules. Conceptually, it resembles prior models such as LMFusion~\citep{shi2024llamafusion}, which augment language models with a separate branch for handling image generation.

\textbf{Training and Evaluation Details.}
All \arch{} variants are trained on 1.2M images from the ImageNet~\cite{imagenet} training split at a resolution of $256\times256$ with a global batch size of 2048 and a constant learning rate of $5e\text{-}4$ for 240K steps. 
Instead of using class IDs for the ImageNet experiments, we use the corresponding natural language captions for both training and evaluation.
For both \baselinecouple{} and \arch{}, we only fine-tune parameters specialized for image generation while keeping parameters for image understanding frozen.
For evaluation, we report FID, Inception Score, Precision, and Recall on ImageNet following previous convention~\cite{peebles2023scalable}.
Additional details can be found in Appendix \ref{app:training_details}.

%% file: tables/gen_results.tex
\begin{table*}[h!]
\centering
\caption{\textbf{Comparison of generative models} across FID, IS, Precision, Recall, parameters, steps, and inference time on ImageNet-50K. For \arch{}, we report the number of parameters activated per timestep, given that it has a timestep-expert architecture where only a subset of layers is used at each step. Rel. Time: inference time relative to \arch{}. $\dagger$: taken from MaskGIT~\citep{chang2022maskgit}\looseness-1}
\vspace{-0.2cm}
\resizebox{1.0\textwidth}{!}{%
\begin{tabular}{c | llrrrrrcc}
\toprule
&\textbf{Model} & \textbf{FID}↓ & \textbf{IS}↑ & \textbf{Pre}↑ & \textbf{Rec}↑ & \textbf{\#Params} & \textbf{\#Step} & \textbf{Time (s / img)} & \textbf{Rel. Time} \\
\midrule
\multirow{5}{0.3cm}{\rotatebox{90}{\textbf{\shortstack{Diffusion\\Models}}}} & ADM~\citep{dhariwal2021diffusion}           & 10.94 & 101.0 & 0.69 & 0.63 & 554M & 250 & 9.677 & 168 \\
& CDM~\citep{ho2022cascaded}           & 4.88  & 158.7 & --   & --   & --   & 8100 & -- \\
 &LDM-4-G~\citep{rombach2022high}       & 3.60  & 247.7 & --   & --   & 400M & 250 & -- \\
& DiT-L/2~\citep{peebles2023scalable}       & 5.02  & 167.2 & 0.75 & 0.57 & 458M & 250 & 1.786 & 31 \\
 &DiT-XL/2~\citep{peebles2023scalable}       & 2.27  & 278.2 & 0.83 & 0.57 & 675M & 250 & 2.592 &45 \\
\midrule
\multirow{2}{*}[1.7ex]{\rotatebox[origin=c]{90}{\parbox{1.2cm}{\centering \textbf{Masked\\Models}}}} & MaskGIT~\citep{chang2022maskgit}       & 6.18  & 182.1 & 0.80 & 0.51 & 227M & 8 & 0.029 &0.5 \rule{0pt}{3ex} \\
 & MAGE~\citep{li2023mage}           & 6.93 & 195.8 & -- & -- & 230M & -- & -- \rule[-1.5ex]{0pt}{3ex} \\
\midrule
 \multirow{5}{0.3cm}{\rotatebox{90}{\textbf{\shortstack{AR\\Models}}}} & VQVAE-2$^\dagger$~\citep{razavi2019generating}      & 31.11 & $\sim$45 & 0.36 & 0.57 & 13.5B & 5120 & -- \\
 &VQGAN$^\dagger$~\citep{esser2021taming}        & 18.65 & 80.4 & 0.78 & 0.26 & 227M & 256 & 1.094 & 19 \\
 &VQGAN~\citep{esser2021taming}       & 15.78 & 74.3 & --   & --   & 1.4B & 256 & 1.382 & 24 \\
 &ViT-VQGAN~\citep{yu2021vector}         & 4.17  & 175.1 & --   & --   & 1.7B & 1024 & 1.382 & 24 \\
 &RQTran.~\citep{lee2022autoregressive}       & 7.55  & 134.0 & --   & --   & 3.8B & 68 & 1.210 & 21 \\
\midrule
 \midrule
 \multirow{4}{0.6cm}{\rotatebox{90}{\textbf{\shortstack{Unified\\Models}}}} &Show-o~\citep{xie2024show} &31.26 &98.7 &0.55 & 0.69 & 1.3B & 50& 2.493 &48 \\
 &Janus Pro~\citep{chen2025janus} & 23.68 &105.2 &0.58 &0.49 & 1.5B & 576 & 0.311 & 6\\
 & \cellcolor{gray!10}\baselinecouple{}  (Ours) &\cellcolor{gray!10}6.33  &\cellcolor{gray!10}192.4 &\cellcolor{gray!10}0.80 &\cellcolor{gray!10}0.67 &\cellcolor{gray!10}2.0B &\cellcolor{gray!10}40 &\cellcolor{gray!10}0.158 &\cellcolor{gray!10}3\\
 & \cellcolor{green!10}\arch{}  (Ours) &\cellcolor{green!10}5.79  &\cellcolor{green!10}213.1 &\cellcolor{green!10}0.78 &\cellcolor{green!10}0.69 & \cellcolor{green!10}500M & \cellcolor{green!10}40 & \cellcolor{green!10}0.052 & \cellcolor{green!10}1 \\
\bottomrule
\end{tabular}}
\label{tab:gen_results}
\end{table*}

%% file: tables/mu_results.tex
\begin{table*}[t!]
    \centering
        \caption{\textbf{Results on comprehensive image understanding benchmarks.} Best scores highlighted in \textbf{bold}. 
        Since our \arch{} is an expert architecture, we report the number of activated parameters used for image understanding. \arch{} preserves Qwen2-VL-2B’s strong understanding performance.}
        \vspace{-0.3cm}
    \begin{subtable}[t]{\textwidth}
        \centering
        \fontsize{7pt}{8pt}\selectfont
        \renewcommand{\arraystretch}{1.1}
        \resizebox{\textwidth}{!}{
        \begin{tabular*}{\linewidth}{lcccccccccc}
        \toprule
         \textbf{Model} & \textbf{MMBench} & \textbf{SEED} & \textbf{POPE} & \textbf{MM-Vet} & \textbf{MME-P} & \textbf{MMMU} & \textbf{RWQA} & \textbf{TEXTVQA} & \textbf{\#Params} & \textbf{TFLOPs} \\
         \midrule
        EMU2 Chat \citep{sun2024generative} & - & 62.8 & - & 48.5 & - & 34.1 & - & 66.6 & 34B & 5.4 \\
        Chameleon \citep{team2024chameleon} & 19.8 & 27.2 & 19.4 & 8.3 & 202.7 & 22.4 & 39.0 & 0.0 & 7B & 3.6 \\
        Chameleon \citep{team2024chameleon} & 32.7 & - & 59.8 & 9.7 & 604.5 & 38.8  & 39.2 & 0.0 & 34B & 17.4 \\
        Seed-X~\citep{ge2024seed} & 70.1 & 66.5 & 84.2 & 43.0 & 1457.0 & 35.6  & - & - & 17B & 11.1 \\
        VILA-U \citep{wu2024vila} & 66.6 & 57.1  & 85.8 & 33.5 & 1401.8 & 32.2  & 46.6 & 48.3 & 7B & 3.6 \\
        EMU3 \citep{wang2024emu3} & 58.5 & 68.2 & 85.2 & 37.2 & 1243.8 & 31.6 & 57.4 & 64.7 & 8B & 4.1 \\
        MetaMorph \citep{tong2024metamorph} & 75.2 & 71.8 & - & - & - &  \textbf{41.8} & 58.3 & 60.5 & 8B & 1.1 \\
        Show-o \citep{xie2024show} & - & - & 80.0 & - & 1097.2 & 27.4 & - & - & 1.3B & 0.7 \\
        Janus \citep{wu2024janus} & 69.4 & 63.7 & 87.0 & 34.3 & 1338.0 & 30.5  & - & - & 1.5B & 0.8\\
        Janus Pro \citep{chen2025janus} & \textbf{75.5} & 68.3 & 86.2 & 39.8 & 1444.0 & 36.3 & - & - & 1.5B & 0.8 \\
        
        Qwen2-VL-2B~\citep{wang2024qwen2} & 74.9 & \textbf{72.4} & \textbf{87.3} & \textbf{51.5} & \textbf{1501.4} & 41.1 & \textbf{60.7} & \textbf{79.7} & 2B & 0.4 \\

        \rowcolor{green!10}\arch{}  & 74.9 & \textbf{72.4} & \textbf{87.3} & \textbf{51.5} & \textbf{1501.4} & 41.1 & \textbf{60.7} & \textbf{79.7} & 2B & 0.4 \\
        \bottomrule
        \end{tabular*}
        }
    \end{subtable}%
     \label{tab:und_results}
\end{table*}

%% file: sections/5_results_discussion.tex
\section{Results and Discussion}
\subsection{Image Generation and Understanding Results}
We evaluate \arch{} on both image generation (Table~\ref{tab:gen_results}) and multimodal understanding (Table~\ref{tab:und_results}) tasks. 
Table~\ref{tab:gen_results} reports quantitative comparison between \arch{}, recent unified models, and leading image generation models. We evaluate each model in terms of generation quality, activated parameters for each inference step, and inference efficiency. 
 All inference times are measured on a single NVIDIA L40 GPU with batch size 50.

\arch{} achieves better FID scores compared to state-of-the-art unified models~\cite{xie2024show, wu2024janus, chen2025janus} that are pretrained on the mixture of ImageNet and other large-scale image-caption datasets, while achieving much faster inference speed, i.e., 48$\times$ faster than Show-o~\cite{xie2024show} and 6$\times$ faster than Janus Pro~\cite{chen2025janus}.  
Moreover, \arch{} outperforms its respective baseline, \baselinecouple{}, which is conceptually similar to  LMFusion~\cite{shi2024llamafusion}, with much fewer activated parameters per flow-matching step and 3$\times$ faster inference speed. The computational cost of \baselinecouple{} is 28.3 TFLOPs per forward pass, compared to only 7.08 TFLOPs for \arch{}, further underscoring the efficiency of the proposed method.\looseness-1

 In addition, \arch{} exhibits competitive performance compared to diffusion models \cite{dhariwal2021diffusion,ho2022cascaded,rombach2022high,peebles2023scalable}, Masked Models~\cite{chang2022maskgit,li2023mage} and Auto-regressive (AR) models~\cite{razavi2019generating, esser2021taming, yu2021vector, lee2022autoregressive} that are specialized for image generation, achieving better parameter and inference-time efficiency. These results suggest \arch{} as a promising, efficient, and effective architecture for image generation. Qualitative results on ImageNet are provided in Appendix~\ref{app:qualitative}.

Table~\ref{tab:und_results} presents results on multimodal understanding benchmarks~\cite{mmbench,seedbench,pope,mmvet,MME,mmmu,textvqa}.
\arch{} achieves competitive or superior performance compared to recent unified models. By effectively leveraging a frozen vision-language backbone, the understanding capability of \arch{} is inherited from its pretrained backbone model Qwen2-VL-2B-Instruct~\citep{wang2024qwen2}, and therefore matches the performance of the backbone itself. 
This approach aligns with concurrent studies~\citep{chen2025blip3ofamilyfullyopen,lin2025uniworld}, which also employ frozen backbones to fully exploit the pretrained understanding strength.\looseness-1

\subsection{Ablation Studies} \label{subsec:abl}

\begin{figure}[!t]
  \centering
  \includegraphics[width=0.9\linewidth]{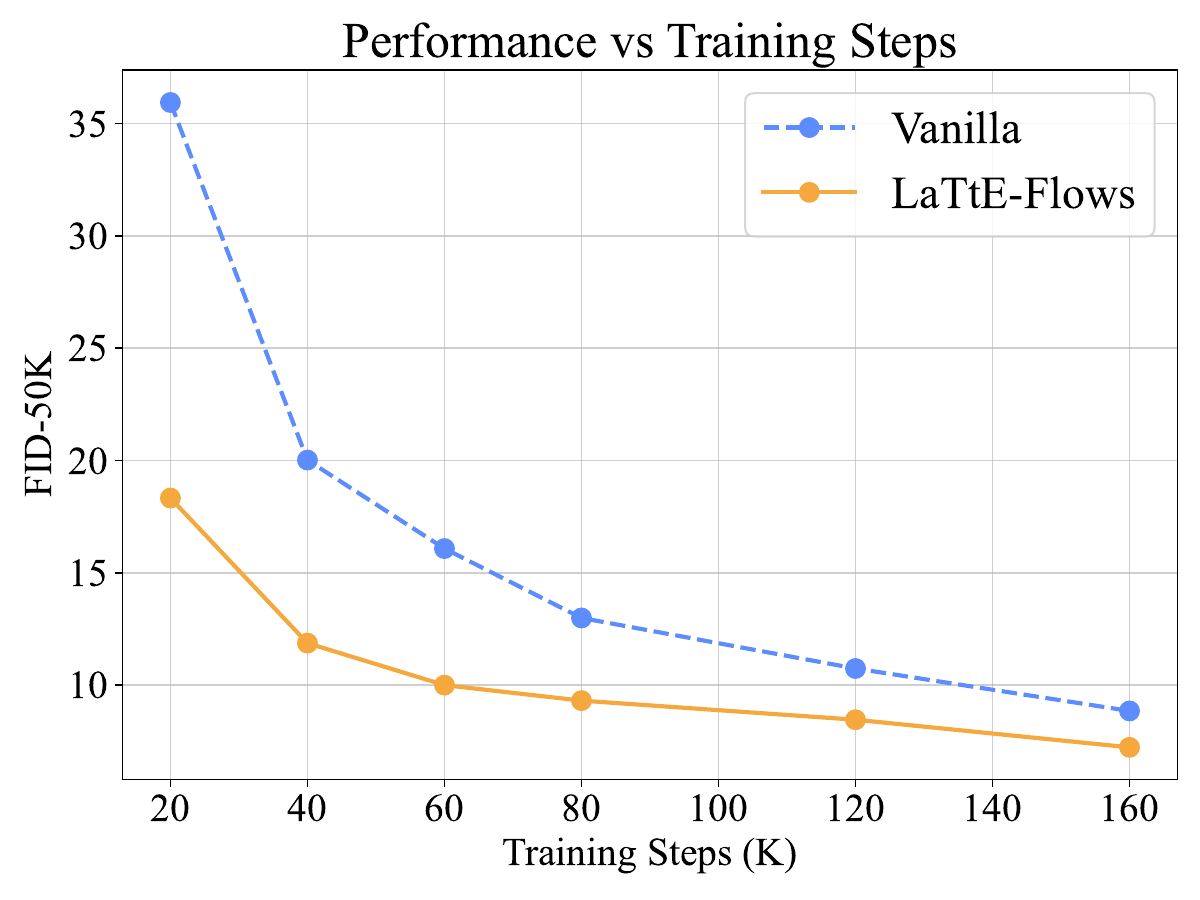}
  \vspace{-0.2cm}
  \caption{\textbf{Training dynamics of \arch{} vs. \baselinecouple{}.} FID on ImageNet 50K.}
  \label{fig:performance_train_step}
\vspace{-0.5cm}
\end{figure}

\paragraph{Faster Convergence Rate of \arch{}.}
Figure~\ref{fig:performance_train_step} illustrates the training dynamics of \arch{} compared to the \baselinecouple{} baseline.
We observe that \arch{} exhibits a significantly faster convergence rate during training, reaching competitive image generation performance (lower FID) in fewer training steps.\looseness-1

We attribute this favorable property of \arch{} to the layerwise timestep-expert architecture. As noted in prior work~\cite{eDiff,HangGLB00GG23}, the slow convergence of diffusion models is partially due to the conflicting optimization directions of different timesteps. Optimizing for timesteps that are close can benefit each other, while optimizing timesteps that are far away can interfere with each other. \arch{}'s layerwise timestep-expert architecture alleviates this challenge by distributing timesteps across different transformer layers. 

\begin{figure}[!t] 
  \centering
  \includegraphics[width=0.36\textwidth]{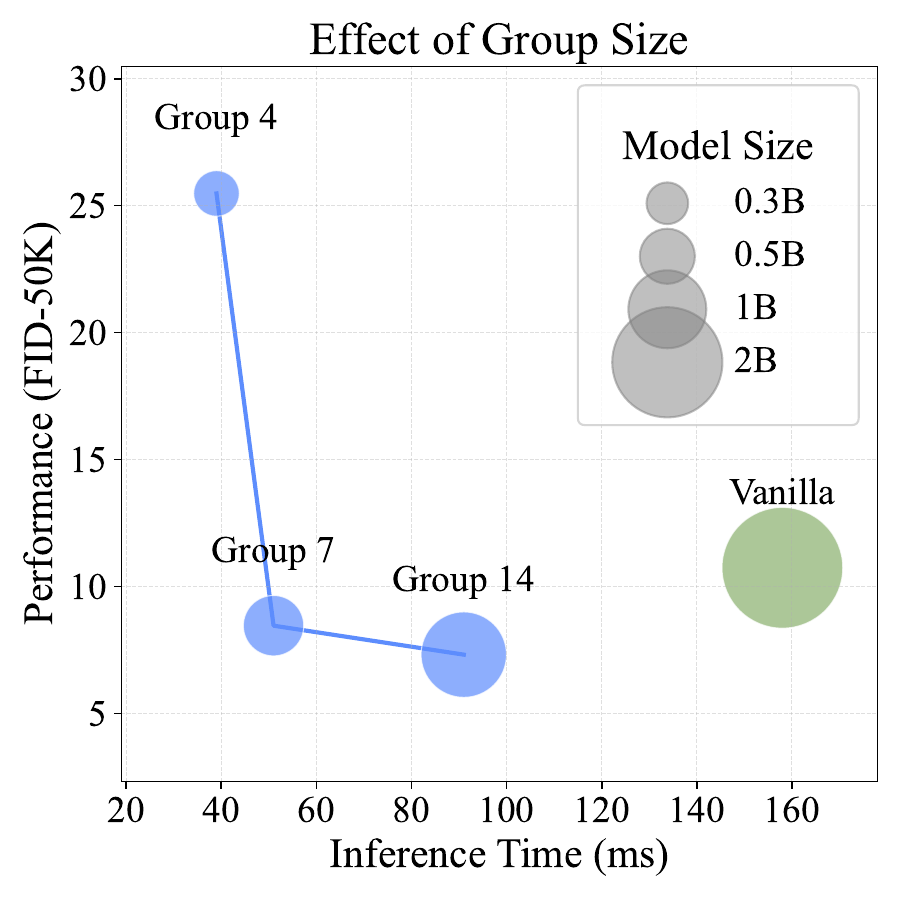}
  \vspace{-0.3cm}
  \caption{\textbf{Effect of group size in \arch{}}. 
  }
  \label{fig:group_size}
  \vspace{-0.3cm}
\end{figure}

\begin{figure*}[!thbp]
	\centering
	\includegraphics[width=0.98\linewidth]{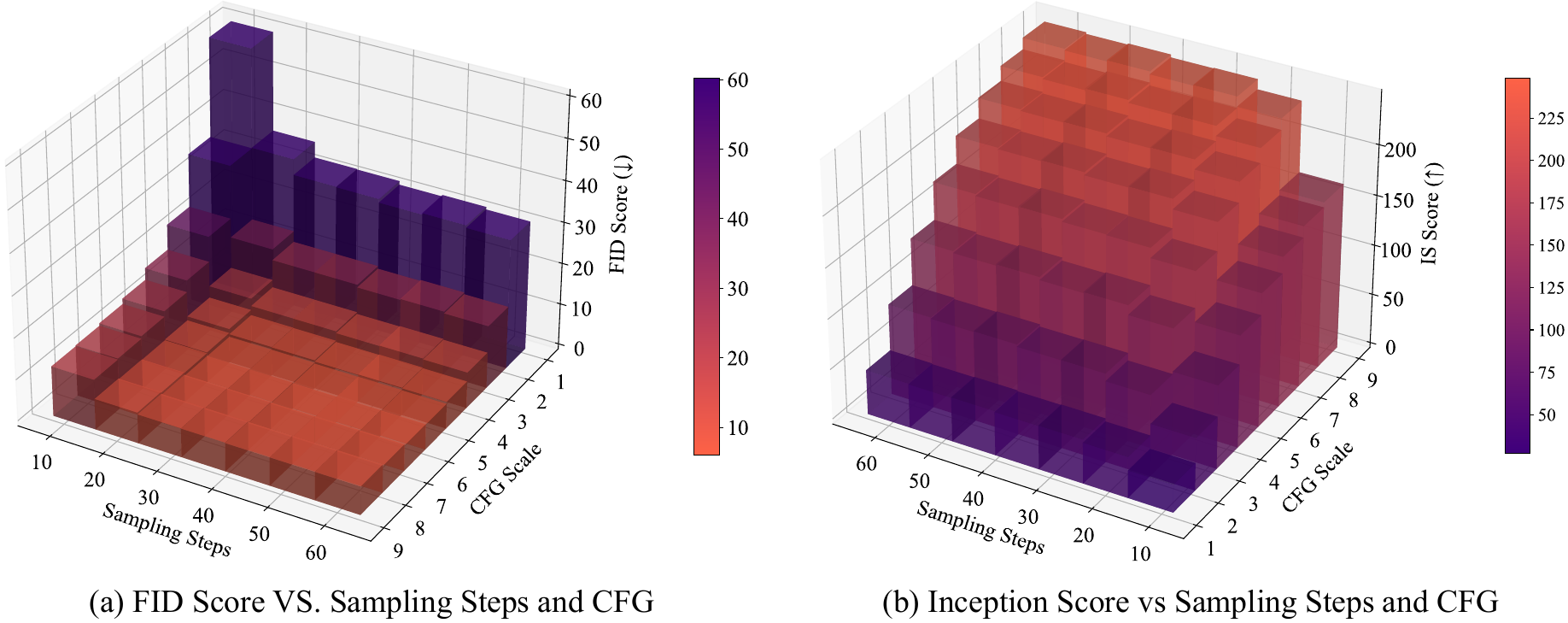}
	\caption{\textbf{Impact of \# sampling steps and CFG strength on Inception Score and FID.}
    }
\label{fig:step_cfg}
\end{figure*}

\paragraph{Impact of Varying Group Size.}
We also investigate how the timestep-expert group size $M$ affects the trade-off between generation quality and inference efficiency. Specifically, we train \arch{} with group sizes $M \in \{4, 7, 14\}$, corresponding to partitioning the transformer layers into 7, 4, and 2 expert groups, respectively.
Figure~\ref{fig:group_size} reports results at 120K training steps. We observe that larger group sizes consistently improve generation quality, as measured by FID, due to increased modeling capacity. However, this comes at the cost of reduced inference speed, since more layers are executed per timestep. 
Both $M\!=\!7$ and $M\!=\!14$ achieve better generation quality and efficiency compared to the baseline \baselinecouple{}, which applies all 28 layers at every step.
Thus, considering the trade-off between performance and efficiency, we select $M\!=\!7$ as the default group size in our main results in Table \ref{tab:gen_results}, which offers strong generation quality with substantial sampling speedups.

\paragraph{Effect of Timestep-Conditioned Residual Attention.}
\input{tables/res_attn}
To quantify the effect of timestep-conditioned residual attention, we compare \arch{} against a variant with the timestep-conditioned residual attention removed. 
As shown in Table~\ref{tab:res_attn}, removing residual attention leads to a notable degradation across multiple metrics,  highlighting the effectiveness of time-conditioned attention across layers. Adding timestep-conditioned residual attention does not introduce additional inference time cost.

\paragraph{Effect of Sampling Steps and CFG.}

Figure \ref{fig:step_cfg} shows the impact of varying the number of sampling steps and classifier-free guidance scale (CFG) on image generation quality.
We observe that increasing the number of steps generally improves image generation quality, leading to lower FID and higher Inception Score. However, as the number of sampling steps surpass 40, performance improvements become marginal.
In general, higher CFG leads to better Inception Score, but for FID, once the CFG goes beyond 5, performance starts to decrease slightly.

\paragraph{Timestep Condition in Residual Attention.}
\label{sec:results_time}

\input{figures/couple_attn_fig}

To better understand the role of timestep conditioning in residual attention, we perform an in-depth analysis on \arch{}. 
Specifically, we first investigate how attention patterns evolve across transformer layers and sampling timesteps in baseline models.
We quantify the sequential similarity between adjacent layers at each timestep using a total variation-based metric:
\begin{equation}
\scalebox{0.85}{$
      S(\bm{A}^l, \bm{A}^{l+1})\!=\!1 - \frac{1}{2} \sum_i \left|\mathrm{Softmax}\left(\bm{A}^{l}_i\right) - \mathrm{Softmax}\left(\bm{A}^{l+1}_i\right)\right|,
      $}
\end{equation}
where $\mathrm{Softmax}\left(\bm{A}^{l}_i\right)$ is the softmax-normalized $i$-th row of attention map $\bm{A}^{l}$.
 Higher values of $S$ reflect greater similarity in image attention maps between successive layers.
 
Figure~\ref{fig:couple_attn} (a) shows how sequential similarity in \baselinecouple{} evolves throughout the sampling process, averaged over 100 randomly selected samples.
We observe that early in sampling, attention maps across layers show low similarity, but as generation progresses, especially in later timesteps, similarity increases, sometimes approaching 1.0 in early layers. This motivates using residual attention for efficient reuse, with dynamic gating needed to adapt to varying similarity patterns across timesteps.

Figure~\ref{fig:couple_attn} (b) shows timestep-conditioned residual attention gates in \arch{}, which modulate how much past-layer attention is reused. As seen across all heads (Figure~\ref{fig:attn_head_couple}), gating remains stable across timesteps within a head but varies between heads, indicating specialization. These results highlight the effectiveness of dynamic, head-specific residual attention in flow-matching generation. 

%% file: tables/res_attn.tex
\begin{table}[!t]
\caption{\textbf{Effect of time-conditioned residual attention.}}
\vspace{-0.3cm}
\centering
\resizebox{0.9\linewidth}{!}{%
\begin{tabular}{llrrrr}
\toprule
\textbf{Model} & \textbf{FID}↓ & \textbf{IS}↑ & \textbf{Pre}↑ & \textbf{Rec}↑ \\
\midrule
\arch{} &5.79  &213.1 & 0.78  & 0.69  \\
- w/o Residual Attention &8.26&157.0&0.75&0.61 \\
 
\bottomrule
\end{tabular}%
}
\label{tab:res_attn}
\vspace{-0.3cm}
\end{table}

%% file: figures/couple_attn_fig.tex
\begin{figure*}[!t]
    \centering
\includegraphics[width=0.99\linewidth]{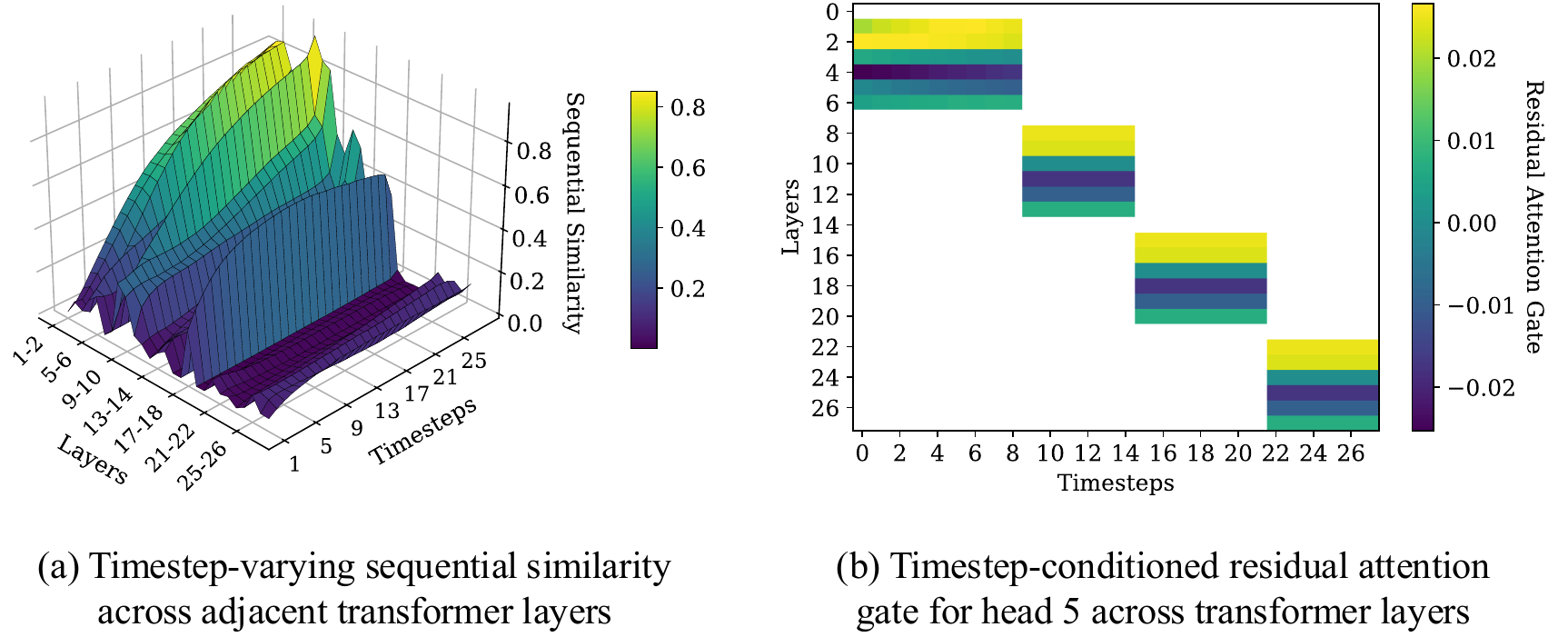}
    \caption{\textbf{Timestep-conditioned residual attention analysis.} (a) Visualization of attention behavior in \baselinecouple{} and (b) learned residual gating patterns in \arch{}.}
    \label{fig:couple_attn}
    \vspace{-0.3cm}
\end{figure*}

%% file: sections/6_conclusion.tex
\section{Conclusion}
In this work, we present \archlong{} (\arch{}), a novel architecture that improves the efficiency of diffusion/flow-based transformer within the unified model setting.
\arch{} introduces two key novel architectural innovations: \textbf{Layerwise Timestep Experts}, which reduces sampling complexity by specializing transformer layers to distinct timestep intervals, and \textbf{Timestep-Conditioned Residual Attention}, which facilitates adaptive reuse and refinement of attention structures across layers. 
Extensive experimental evaluations demonstrate that \arch{} not only achieves strong multimodal understanding and image generation performance, but also achieves around 6× faster inference compared to existing unified models.\looseness-1

%% file: sections/X_sup.tex
\section{\arch{} \blend{}}
\label{app:blend}

\input{figures/overview_all}

To demonstrate that \arch{} is not tied to a specific flow-matching architecture, we also introduce \textbf{\arch{} \blend{}} and apply our method on the \blend{} architecture as well.
Figure~\ref{fig:diff_vlm_arch_all} shows that \arch{} \blend{} unifies the image generation and understanding components through a partially shared transformer layer.
Here, each layer consists of task-specific submodules with separate parameters for generation and understanding, and a set of shared submodules that are used by both tasks.  
This design enables tighter fusion between generation and understanding signals, facilitating more effective information exchange while maintaining flexibility to specialize for each modality.

We also construct the baseline model \baselineblend{}, which matches the architectures of \arch{} \blend{}, but excludes both the Layerwise Timestep Experts and Timestep-Conditioned Residual Attention mechanisms, allowing us to directly evaluate the effectiveness of these proposed mechanisms on a different architecture.
The \baselineblend{} baseline unifies generation and understanding computations within shared layers, akin to the design of Transfusion~\citep{zhou2024transfusion}. We perform a full parameter fine-tuning for \baselineblend{} and \arch{} \blend{}.

Table~\ref{tab:full_comparison} reports quantitative comparison between \baselineblend{}, \arch{} \blend{}, recent unified models, and leading image generation models. 
 We show that both \arch{} variants outperform their respective baselines, \baselineblend{} and \baselinecouple{}, which are conceptually similar to Transfusion~\cite{zhou2024transfusion} and LMFusion~\cite{shi2024llamafusion}, with much fewer activated parameters per flow-matching step and 3 to 4$\times$ faster inference speed.

\input{tables/gen_results_all}

\section{\arch{} Attention Module}
\label{app:latte_attn}

Figure \ref{fig:latte_flow_attn} illustrates the architecture of the \arch{} Attention module.
Our framework applies 3D Rotary Positional Embeddings (RoPE)~\citep{su2024roformer} from the pretrained VLM to multimodal hidden states and uses new 2D Rotary Positional Embeddings to the generative image tokens. We adopt bi-directional attention on generative image tokens, and all generative image tokens are allowed to attend to previous multimodal tokens.

\input{figures/latte-flow_attn_fig}

\section{Implmentation Details}
\label{app:imp_details}

\paragraph{Timestep Distribution.}
\label{app:timestep}
To enable Layerwise Timestep Experts, \arch{} partitions the model into $K\!=\!4$ non-overlapping layer groups, each containing $M\!=\!7$ consecutive layers for the final results. These groups are designed to operate over distinct intervals of the flow-matching timesteps.
During training, we use $T\!=\!1000$ flow-matching steps, which are initially divided uniformly into four intervals: $[1000.0, 750.25]$, $[750.25, 500.50]$, $[500.50, 250.75]$, and $[250.75, 0]$. 
To encourage robustness near interval boundaries and promote smooth transitions across groups, we introduce a 100-step overlap between adjacent timestep intervals during training. This overlap allows boundary timesteps to be seen by multiple layer groups, improving generalization.
Specifically, layers $1$ through $7$ are assigned to the timestep interval $[1000, 700]$, layers $8$ through $14$ cover $[700, 450]$, layers $15$ through $21$ operate on $[450, 200]$, and layers $22$ through $28$ handle the final interval $[200, 0]$. 
Each group is trained exclusively on its assigned range according to Eq. (\ref{eq:layerwise}), enabling it to specialize in the velocity prediction of that particular segment of the flow-matching timestep interval.\looseness-1

At inference time, we disable overlaps to maintain strict partitioning of timestep intervals.
Consequently, at each denoising step, only the corresponding expert layer group is activated, requiring just $M\!=\!7$ layers per inference step. This contrasts favorably with standard diffusion or flow-matching models that activate all $L\!=\!28$ layers at every step, significantly enhancing generation efficiency.

\paragraph{Training and Evaluation Details.}
\label{app:training_details}
We train all model variants on eight H200 GPUs for approximately four days.
During training, following previous approaches, we employ classifier-free guidance~\citep{ho2021classifier} to guide the sampling process for better sampling quality by amplifying the difference between conditional and unconditional generation with the guidance scale $> 1$. During training, we randomly drop the multimodal condition with $p=10\%$ to facilitate unconditional prediction.

For evaluation, each model generates 50 images for each of 1,000 classes in ImageNet with 40 sampling steps and classifier-free guidance (CFG) of 5 based on our ablation study in Section~\ref{subsec:abl}. We report FID and Inception Score of 50K generated images against 50K real images from the ImageNet validation split. Following previous convention~\cite{peebles2023scalable}, we compute Precision and Recall using 1,000 generated images. All scores are calculated using standard implementations from torch-fidelity~\footnote{\url{https://github.com/toshas/torch-fidelity}}.

\section{User Study}

To complement the automated metrics and further assess the generative quality of \arch{}, we conduct a human preference study comparing our model against two recent unified model baselines, Janus Pro~\citep{chen2025janus} and Show-o~\citep{xie2024show}.
We randomly sample 50 class prompts from ImageNet and generate images for each prompt using all three models. For each prompt, we present the three corresponding images to human evaluators in randomized order to avoid positional bias. We recruit 10 annotators and instruct them to select the image they prefer, with explicit guidance to evaluate along two axes: (1) \emph{photo-realism}, and (2) \emph{semantic accuracy} with respect to the prompt. The full annotation guideline is:
\begin{verbatim}
    Please follow the instructions below when evaluating images:
    Please do not rely solely on overall image aesthetic quality (e.g., 
    style, beauty, artistic appeal) when determining preference. You should 
    also pay attention to photo-realism, as ImageNet-1k consists of photo-
    realistic images.
    
    In addition, a model may generate a visually impressive image that is 
    semantically incorrect. Please carefully verify that the main object or 
    animal in the image matches the caption. Check for correct species and 
    object identity as described on the left.
    
    Your evaluation should be based primarily on:
    1. Photo-realism
    2. Semantic accuracy (whether the visual content truly corresponds to 
    the caption)
    
    For each row in the table, you will see three images generated by 
    different models for the same caption.
    Please rank the images (1 = best, 3 = worst).
    
    You may assign ties if multiple images are equally good or equally bad.
    For example: 1, 1, 2 → two best images tie for rank 1.
\end{verbatim}

\begin{figure}
    \centering
    \includegraphics[width=0.95\linewidth]{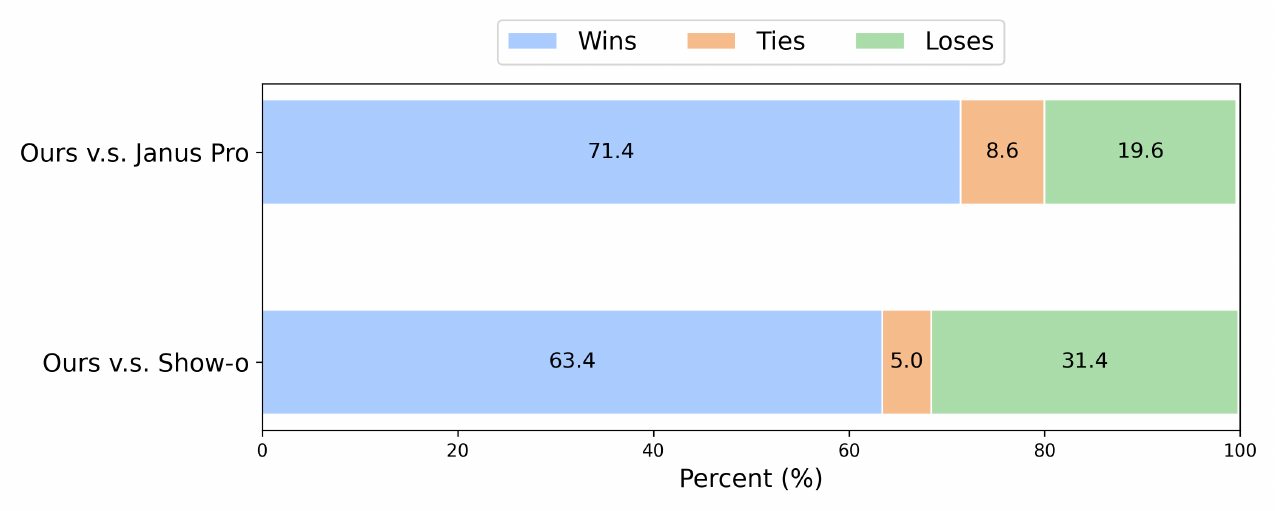}
    \vspace{-0.3cm}
    \caption{\textbf{Human preference study results.} We report pairwise win/tie/loss rates between \arch{} and each baseline.}
    \label{fig:rank}
\end{figure}

\begin{figure}[t!]
    \centering
    \includegraphics[width=0.99\linewidth]{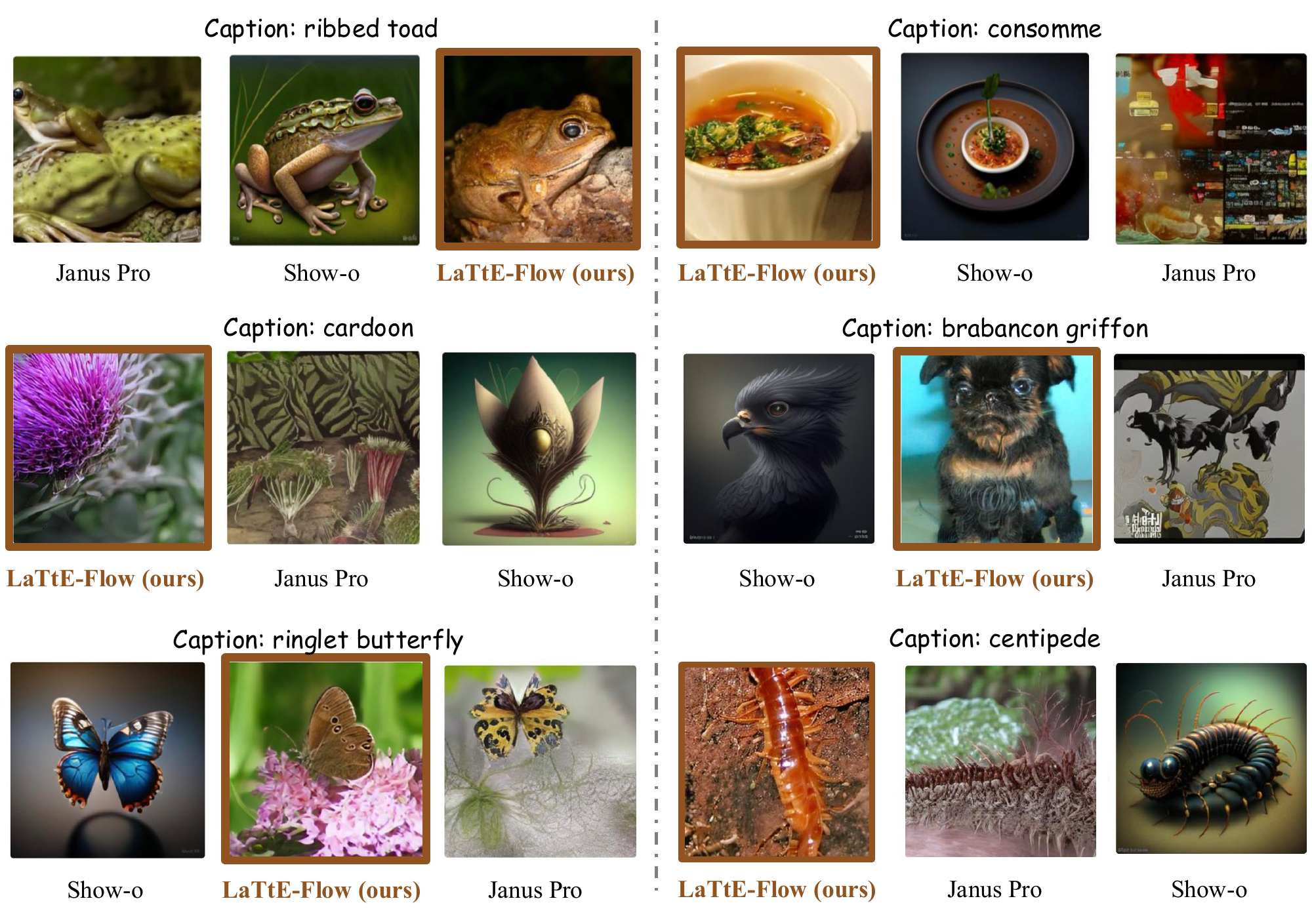}
    \vspace{-0.3cm}
    \caption{\textbf{Qualitative examples of user-study comparisons.} For visualization purposes, we display the model names below each image and highlight the output of \arch{} using a brown frame. Note that in the actual user study, all generated images were anonymized and unframed to avoid revealing model identity or introducing positional bias.}
    \label{fig:user_cases}
\end{figure}

Figure \ref{fig:rank} reports preference rates against baselines. \arch{} is preferred to Janus Pro in 71.4\% of cases (with 8.6\% ties and 19.6\% losses) and preferred to Show-o in 63.4\% of cases (with 5.0\% ties and 31.4\% losses).

Moreover, Figure~\ref{fig:user_cases} presents several qualitative comparison examples used in the study. As shown, Show-o sometimes produces visually appealing images but fails to align with the given prompt. 
Janus Pro, on the other hand, tends to generate images in which the target object loses structural integrity. In contrast, \arch{} is able to produce images that are both photo-realistic and semantically faithful to the prompt.

\section{Qualitative Results}
\label{app:qualitative}

Figure \ref{fig:qulitative} shows qualitative results for sampled 256 $\times$ 256 images generated by \arch{}, conditioned on natural language captions from ImageNet. 
The samples cover a diverse set of object categories and scene types, including animals, plants, vehicles, landmarks, and natural landscapes. 
These qualitative examples further indicate that LaTtE-Flow preserves strong generation quality while using a more efficient layerwise timestep-expert architecture.

\input{figures/qualitative}

\section{Timestep-Conditioned Residual Attention}
\label{app:attn}

Following the experimental setup in Section \ref{sec:results_time}, we also perform an in-depth analysis on the \arch{} \blend{} variant. Figure \ref{fig:blend_attn} (a) shows how this sequential similarity across adjacent layers evolves over the sampling timesteps. The plot shows the mean similarity computed across 100 randomly sampled examples.
We observe that for most of the adjacent layers, the sequential similarity is relatively low at early timesteps, and gradually increases as the timestep progresses, particularly in early layers, where the similarity rises and approaches 1.0. However, the observed similarity pattern varies significantly across timesteps and layers, motivating the need for a timestep-conditioned gating strategy of residual attention flows.\looseness-1

In Figure~\ref{fig:blend_attn} (b), we visualize the learned residual attention gating values for head 11 within \arch{} \blend{}. 
These gates are dynamically modulated by timestep embeddings and control the degree to which residual attention from the previous layer is incorporated into the current layer’s computation. 
To further understand the role of residual attention across heads, Figure \ref{fig:attn_head_blend} displays the gating values for all 12 heads in \arch{} \blend{}. We observe that gating remains relatively stable across timesteps within a specific head, but the patterns differ notably among different heads.
A similar trend is also observed in the \arch{} variant (Figure~\ref{fig:attn_head_couple}), where head-specific gating patterns reflect different behaviors.
In summary, these results validate the design of timestep-conditioned, head-specific residual attention. The gating mechanism enables adaptive reuse of earlier attention.

\input{figures/blend_attn_fig}
\input{figures/couple_attn_heads}
\input{figures/blend_attn_heads}

\section{The Use of Large Language Models}
In preparing this manuscript, we mainly used large language models (LLMs) as an auxiliary tool for polishing the writing. Specifically, the models were employed to improve sentence fluency, correct grammar errors, and refine clarity of expression. They were not involved in research ideation, experimental design, analysis, or substantive content generation.

\input{sections/7_limitation}

%% file: figures/overview_all.tex
\begin{figure}[!h]
	\centering
	\includegraphics[width=0.99\linewidth]{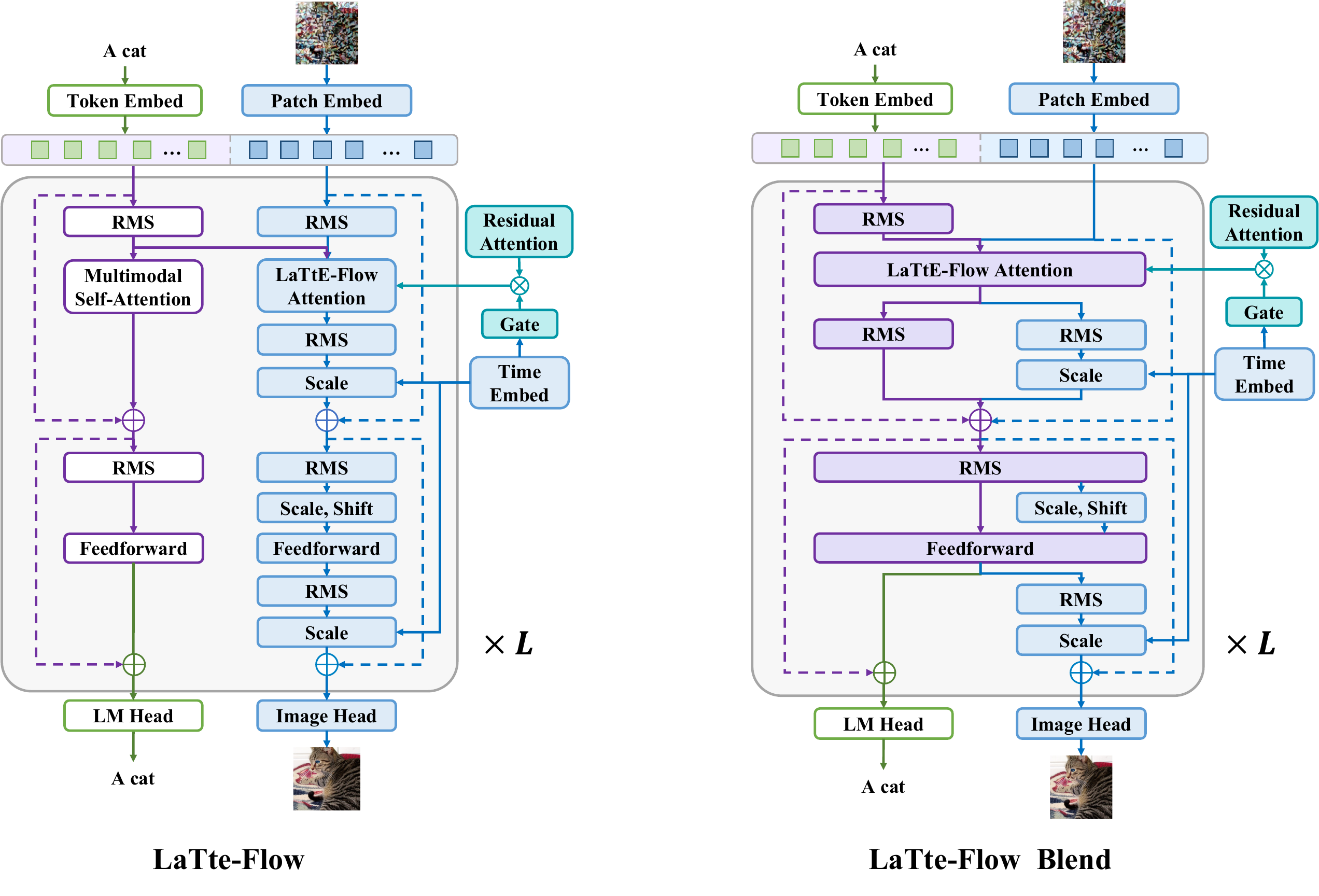}
    \vspace{-0.5cm}
	\caption{\textbf{\archcolor{} overall architecture}. 
    }
\label{fig:diff_vlm_arch_all}
\end{figure}

%% file: tables/gen_results_all.tex
\begin{table}[h!]
\centering
\caption{\textbf{Comparison of generative models} across FID, IS, Precision, Recall, parameters, steps, and inference time on ImageNet-50K. For \arch{}, we report the number of parameters activated per timestep, given that it has a timestep-expert architecture where only a subset of layers is used at each step. Rel. Time: inference time relative to \arch{}. $\dagger$: taken from MaskGIT~\citep{chang2022maskgit}\looseness-1}
\vspace{-0.2cm}
\resizebox{1.0\textwidth}{!}{%
\begin{tabular}{c | llrrrrrcc}
\toprule
&\textbf{Model} & \textbf{FID}↓ & \textbf{IS}↑ & \textbf{Pre}↑ & \textbf{Rec}↑ & \textbf{\#Params} & \textbf{\#Step} & \textbf{Time (s / img)} & \textbf{Rel. Time} \\
\midrule
\multirow{5}{0.3cm}{\rotatebox{90}{\textbf{\shortstack{Diffusion\\Models}}}} & ADM~\citep{dhariwal2021diffusion}           & 10.94 & 101.0 & 0.69 & 0.63 & 554M & 250 & 9.677 & 168 \\
& CDM~\citep{ho2022cascaded}           & 4.88  & 158.7 & --   & --   & --   & 8100 & -- \\
 &LDM-4-G~\citep{rombach2022high}       & 3.60  & 247.7 & --   & --   & 400M & 250 & -- \\
& DiT-L/2~\citep{peebles2023scalable}       & 5.02  & 167.2 & 0.75 & 0.57 & 458M & 250 & 1.786 & 31 \\
 &DiT-XL/2~\citep{peebles2023scalable}       & 2.27  & 278.2 & 0.83 & 0.57 & 675M & 250 & 2.592 &45 \\
\midrule
\multirow{2}{*}[1.7ex]{\rotatebox[origin=c]{90}{\parbox{1.2cm}{\centering \textbf{Masked\\Models}}}} & MaskGIT~\citep{chang2022maskgit}       & 6.18  & 182.1 & 0.80 & 0.51 & 227M & 8 & 0.029 &0.5 \rule{0pt}{3ex} \\
 & MAGE~\citep{li2023mage}           & 6.93 & 195.8 & -- & -- & 230M & -- & -- \rule[-1.5ex]{0pt}{3ex} \\
\midrule
 \multirow{5}{0.3cm}{\rotatebox{90}{\textbf{\shortstack{AR\\Models}}}} & VQVAE-2$^\dagger$~\citep{razavi2019generating}      & 31.11 & $\sim$45 & 0.36 & 0.57 & 13.5B & 5120 & -- \\
 &VQGAN$^\dagger$~\citep{esser2021taming}        & 18.65 & 80.4 & 0.78 & 0.26 & 227M & 256 & 1.094 & 19 \\
 &VQGAN~\citep{esser2021taming}       & 15.78 & 74.3 & --   & --   & 1.4B & 256 & 1.382 & 24 \\
 &ViT-VQGAN~\citep{yu2021vector}         & 4.17  & 175.1 & --   & --   & 1.7B & 1024 & 1.382 & 24 \\
 &RQTran.~\citep{lee2022autoregressive}       & 7.55  & 134.0 & --   & --   & 3.8B & 68 & 1.210 & 21 \\
\midrule
 \midrule
 \multirow{4}{0.6cm}{\rotatebox{90}{\textbf{\shortstack{Unified\\Models}}}} &Show-o~\citep{xie2024show} &31.26 &98.7 &0.55 & 0.69 & 1.3B & 50& 2.493 &48 \\
 &Janus Pro~\citep{chen2025janus} & 23.68 &105.2 &0.58 &0.49 & 1.5B & 576 & 0.311 & 6\\
 &\cellcolor{gray!10}\baselineblend{} (Ours) &\cellcolor{gray!10}6.12 &\cellcolor{gray!10}193.7 &\cellcolor{gray!10}0.78 &\cellcolor{gray!10}0.69 &\cellcolor{gray!10}2.0B &\cellcolor{gray!10}40 &\cellcolor{gray!10}0.185 &\cellcolor{gray!10}4 \\
 & \cellcolor{green!10}\arch{} Blend  (Ours) &\cellcolor{green!10}6.03  &\cellcolor{green!10}193.9 &\cellcolor{green!10}0.77 &\cellcolor{green!10}0.68 &\cellcolor{green!10}500M &\cellcolor{green!10}40 &\cellcolor{green!10}0.061 &\cellcolor{green!10}1 \\
 & \cellcolor{gray!10}\baselinecouple{}  (Ours) &\cellcolor{gray!10}6.33  &\cellcolor{gray!10}192.4 &\cellcolor{gray!10}0.80 &\cellcolor{gray!10}0.67 &\cellcolor{gray!10}2.0B &\cellcolor{gray!10}40 &\cellcolor{gray!10}0.158 &\cellcolor{gray!10}3\\
 & \cellcolor{green!10}\arch{}  (Ours) &\cellcolor{green!10}5.79  &\cellcolor{green!10}213.1 &\cellcolor{green!10}0.78 &\cellcolor{green!10}0.69 & \cellcolor{green!10}500M & \cellcolor{green!10}40 & \cellcolor{green!10}0.052 & \cellcolor{green!10}1 \\
\bottomrule
\end{tabular}}
\label{tab:full_comparison}
\end{table}

%% file: figures/latte-flow_attn_fig.tex
\begin{figure}[!h]
	\centering
	\includegraphics[width=0.6\linewidth]{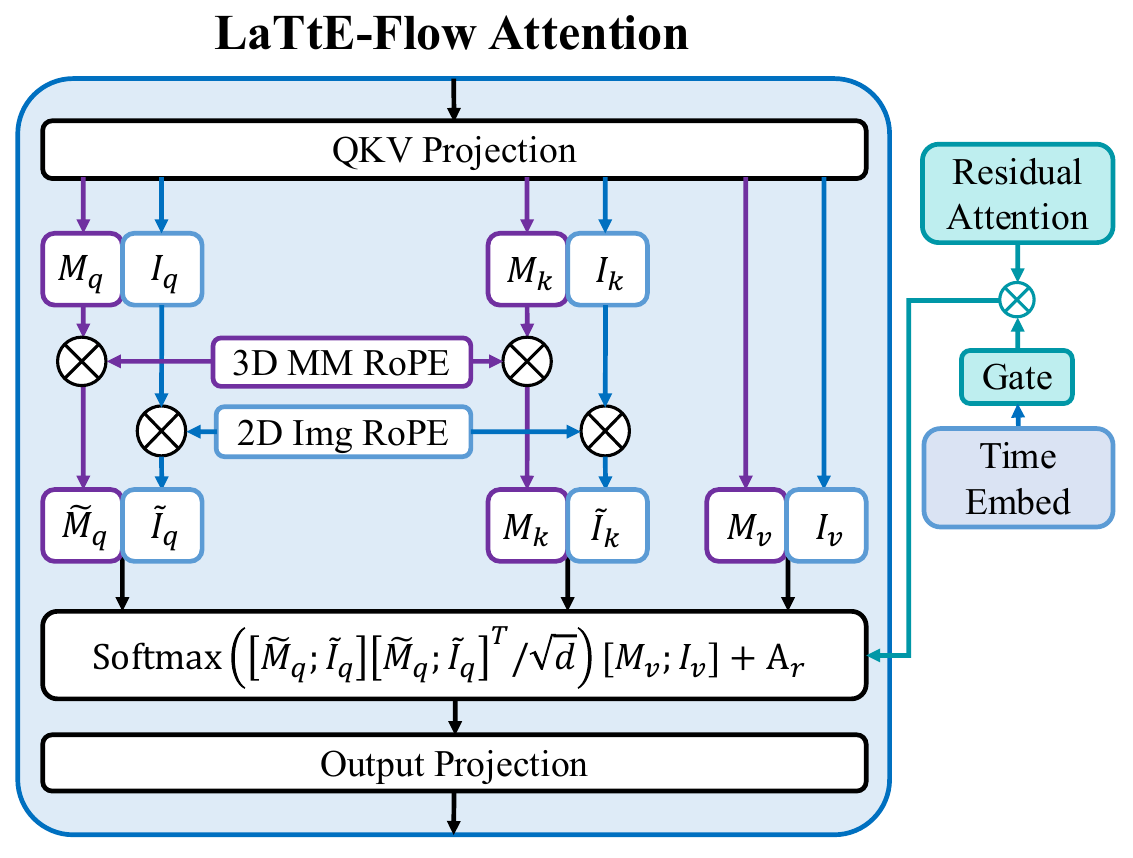}
	\caption{\textbf{\arch{} Attention}}
\label{fig:latte_flow_attn}
\end{figure}

%% file: figures/qualitative.tex
\begin{figure}[!t]
	\centering
	\includegraphics[width=\linewidth]{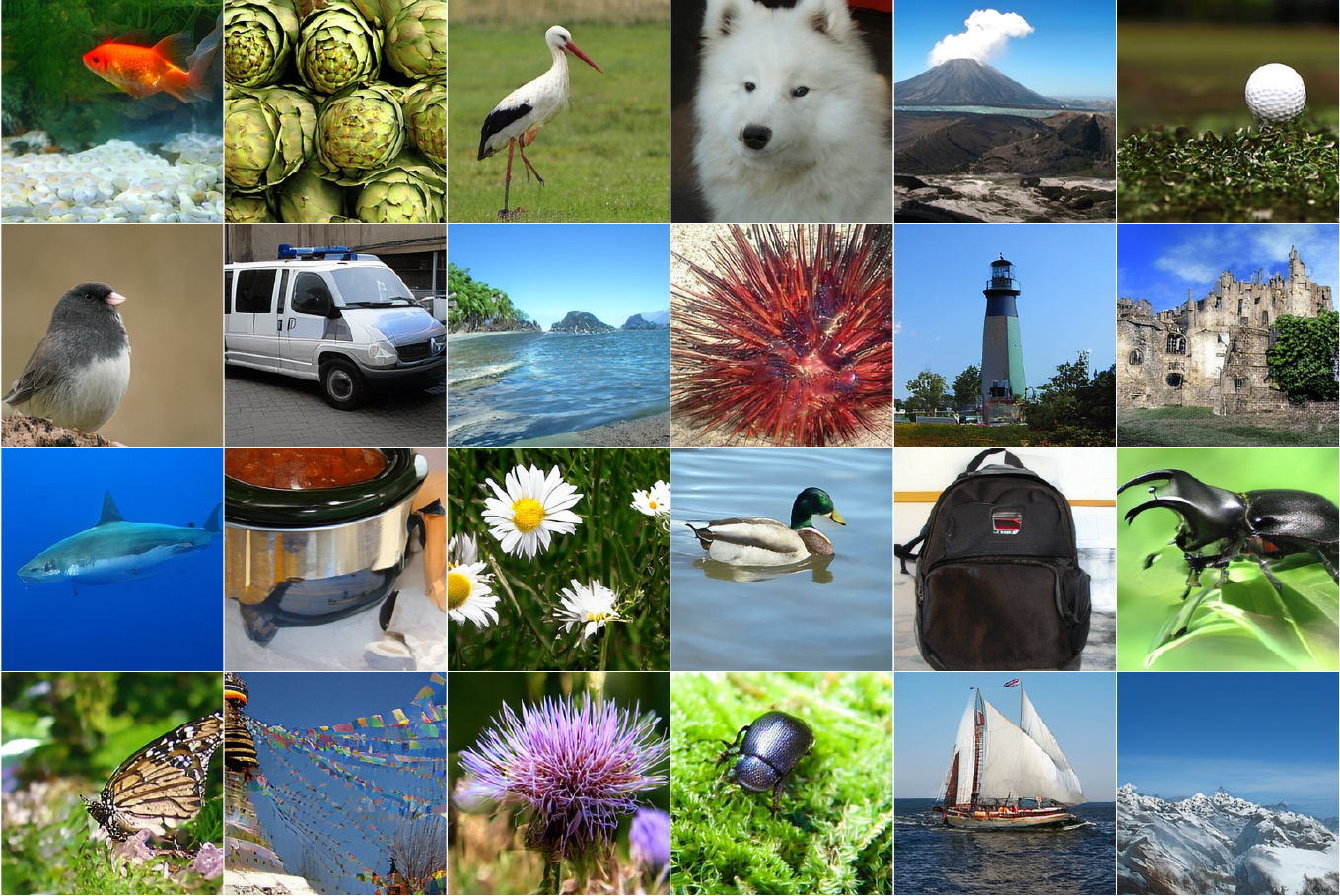}
	\caption{\textbf{Generated 256$\times$256 samples by \arch{} \couple{} trained on ImageNet.}}
\label{fig:qulitative}
\end{figure}

%% file: figures/blend_attn_fig.tex
\begin{figure}[h!]
    \centering
\includegraphics[width=0.99\linewidth]{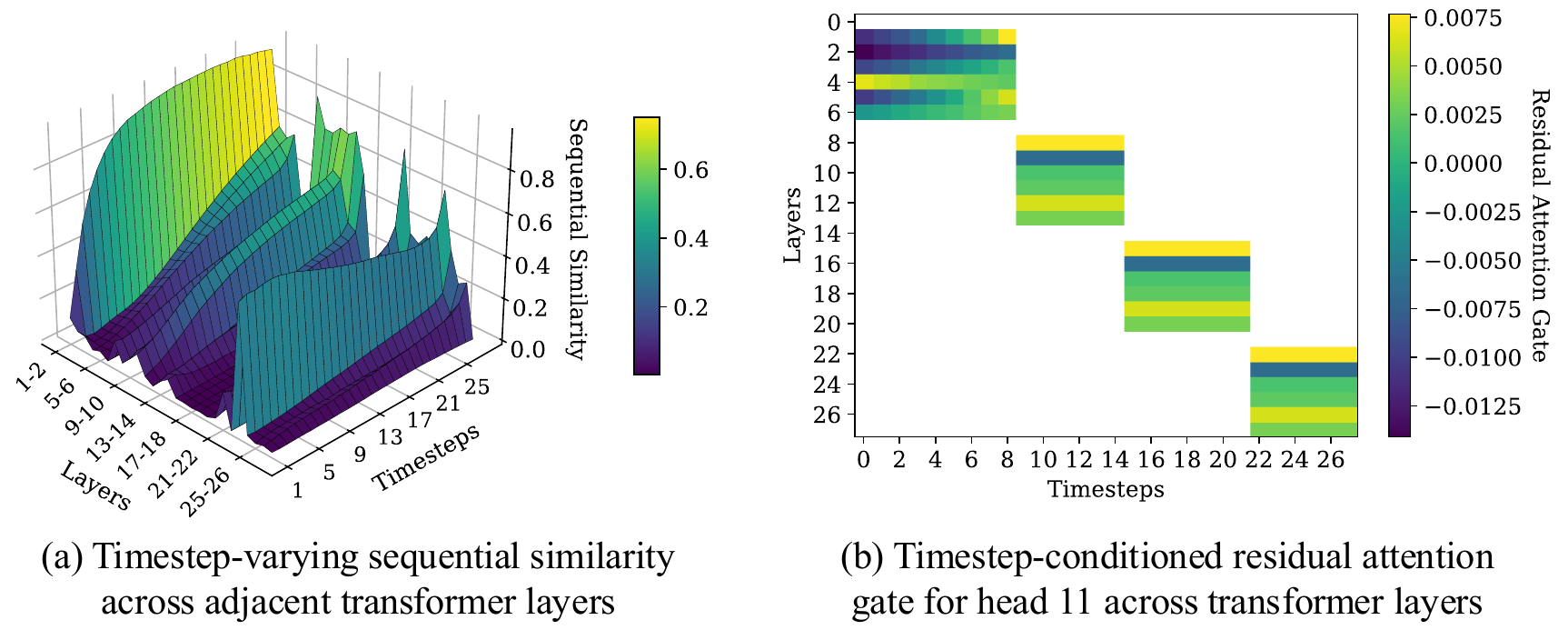}
    \caption{\textbf{Visualization of attention in Baseline Blend and \arch{} Blend.} (a) Sequential similarity between adjacent layers increases over timesteps, particularly in early layers.
(b) Residual attention gating in \arch{} Blend (head 11) shows relatively consistent gating values across timesteps within the same head.
}
    \label{fig:blend_attn}
    \vspace{-0.3cm}
\end{figure}

%% file: figures/couple_attn_heads.tex
\begin{figure*}[h!]
	\centering
	\includegraphics[width=0.97\linewidth]{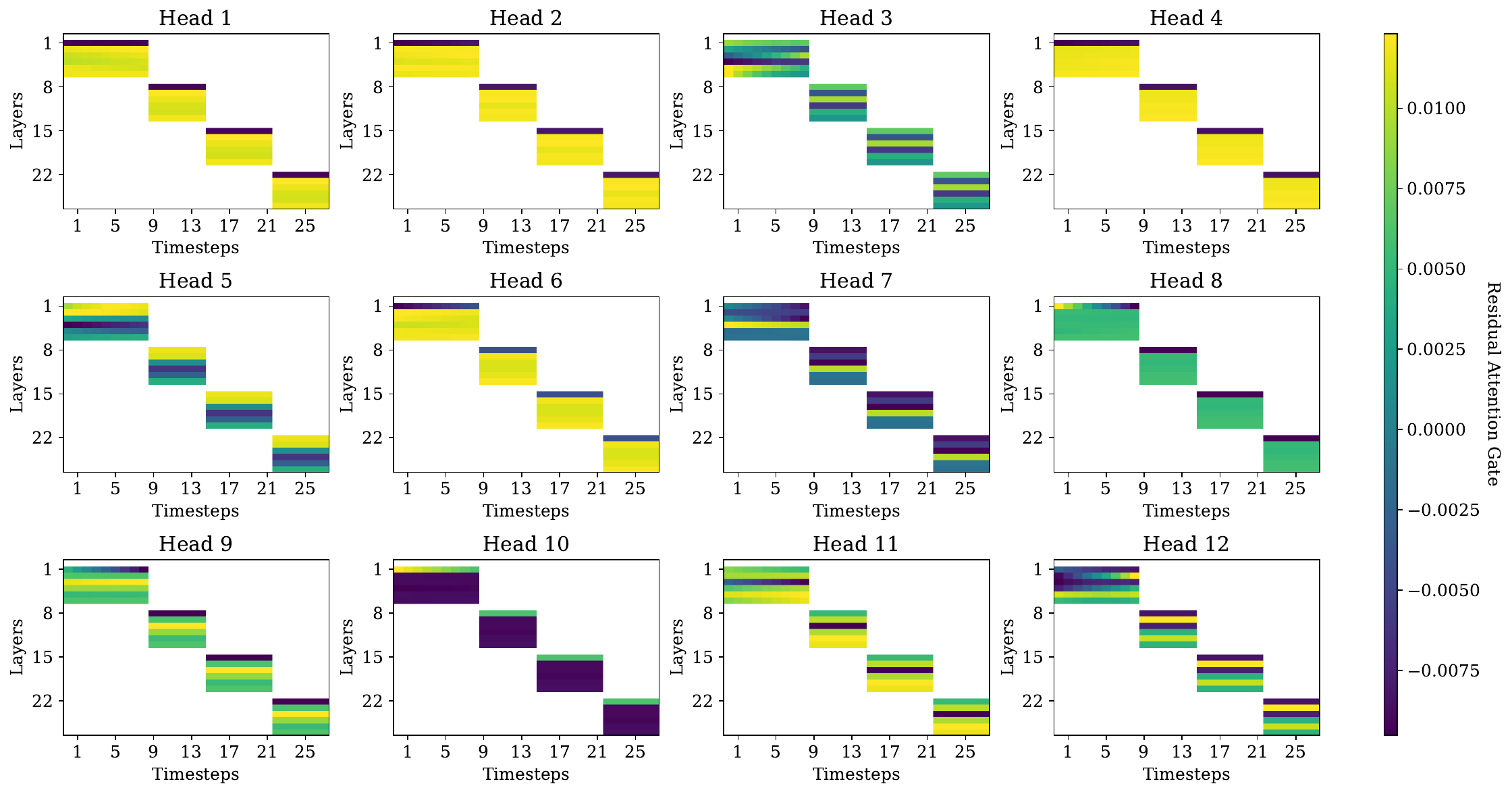}
	\caption{\textbf{Timestep-conditioned residual attention gates across transformer layers in \arch{}.} White regions indicate positions without gating values since residual attention is applied only within predefined layer groups.
Notably, different heads exhibit distinct gating dynamics, with some emphasizing earlier timesteps, while others modulate more strongly in later layers, suggesting head-specific specialization in residual attention.}
\label{fig:attn_head_couple}
    \vspace{-0.4cm}
\end{figure*}

%% file: figures/blend_attn_heads.tex
\begin{figure*}[h!]
	\centering
	\includegraphics[width=0.97\linewidth]{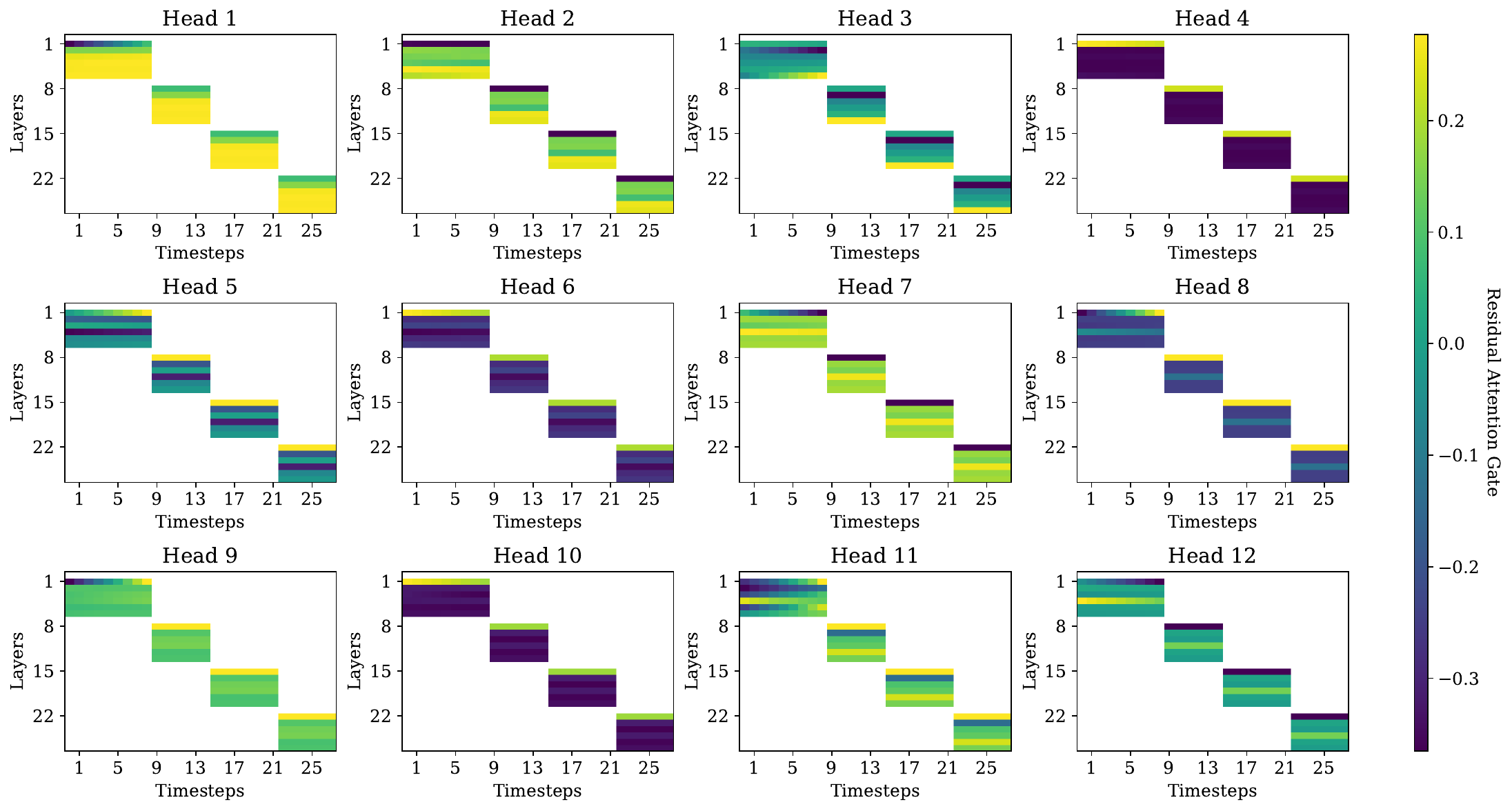}
	\caption{\textbf{Timestep-conditioned residual attention gates across transformer layers in \arch{} \blend{}.} White regions indicate positions without gating values since residual attention is applied only within predefined layer groups.
Notably, different heads exhibit distinct gating dynamics, with some emphasizing earlier timesteps, while others modulate more strongly in later layers, suggesting head-specific specialization in residual attention.}
\label{fig:attn_head_blend}
\end{figure*}

%% file: sections/7_limitation.tex
\section{Limitations}
\label{sec:limitation}

Although \arch{} achieves substantial improvements in sampling efficiency with strong results in multimodal understanding and generation tasks, several limitations remain. First, our experiments involved training \arch{} for only 240K optimization steps, significantly fewer than existing unified multimodal models. Extending the training duration could potentially enhance the model's performance further.
Second, while our uniform timestep distribution with overlapping intervals proved effective, the optimal timestep distributions or layer partitioning strategies remain an open problem. Future work should systematically explore and optimize timestep partitioning strategies.